\documentclass[twoside,11pt]{article}

\usepackage{jmlr2e}
\usepackage{xcolor}
\usepackage{makecell}
\usepackage{array}

\ShortHeadings{Tunnel effects in cognition}{Cornu\'ejols et al.}
\firstpageno{1}

\begin{document}

\title{Tunnel effects in cognition: \\
A new mechanism for scientific discovery and education
}

\author{\name  Antoine Cornu\'ejols \email antoine.cornuejols@agroparistech.fr \\
       \addr MMIP\\
       AgroParisTech, dept. MMIP and INRA UMR-518\\
       16, rue Claude Bernard, F-75231 Paris Cedex, France
       \AND
       \name Andr\'ee Tiberghien \email Andree.Tiberghien@univ-lyon2.fr \\
       \addr Laboratoire ICAR UMR 5191 (Interactions, Corpus, Apprentissage, Repr\'esentations) \\
	Ecole Normale Sup\'erieure de Lyon \\
	15 parvis Ren\'e Descartes – BP 7000 \\
	69342 LYON Cedex 07, France
	\AND
       \name G\'erard Collet 
	}

\editor{}

\maketitle

\begin{abstract}

It is quite exceptional, if it ever happens, that a new conceptual domain be built from scratch. Usually, it is developed and mastered in interaction, both positive and negative, with other more operational existing domains. Few reasoning mechanisms have been proposed to account for the interplay of different conceptual domains and the transfer of information from one to another. Analogical reasoning is one, blending \citep{fauconnier1998conceptual} is another. This paper presents a new mechanism, called 'tunnel effect', that may explain, in part, how scientists and students reason while constructing a new conceptual domain. One experimental study with high school students and analyses from the history of science, particularly about the birth of classical thermodynamics, provide evidence and illustrate this mechanism. The knowledge organization, processes and conditions for its appearance are detailed and put into the perspective of a computational model. Specifically, we put forward the hypothesis that two levels of knowledge, notional and conceptual, cooperate in the scientific discovery process when a new conceptual domain is being built. The type of conceptual learning that can be associated with tunnel effect is discussed and a thorough comparison is made with analogical reasoning in order to underline the main features of the new proposed mechanism.

\end{abstract}


\section{Introduction}

Conceptual domains are tools that bring both a filter and a magnifying glass on our universe. They select and organize. They segment and predict. They allow to describe, to make predictions and to explain our environment. As such, when they are mature, they offer an operational way of tackling the world. Scientific discovery as well as education is concerned with learning conceptual domains. By ``conceptual domain'', we refer to a vast organization of knowledge, such as the knowledge of Newtonian mechanics, or of the Darwinian theory of evolution. 

A conceptual domain has a basic structure of entities and relations at a high level of generality — for example the entity \textit{particle} in physics refers to an abstract concept around which revolves a whole lot of roles (e.g. state, trajectory, and so on) as well as a whole framework of expectations (e.g. a state is completely determinable), theories (e.g. Hamiltonian formulation) and inference and control procedures (e.g. systematic recourse to the least action principle). 

Surprisingly few works have directly dealt with how these conceptual domains come to existence, how they are learned, how they develop and how they adapt. 

Philosophers have chosen to make a distinction between the context of creation and the context of justification of a theory, focusing almost exclusively on this last problem (see \citep{popper1959logic,popper1962conjectures}). Cognitive scientists on the whole, and machine learning scientists in particular, have mostly centered their work on induction of “simple” and isolated concepts in rather poorly structured hypotheses spaces. This is called the data-driven approach to scientific discovery (see \citep{langley1987scientific}). While the results obtained have been spectacular, in many respects they do not bear on the learning of complex conceptual domains and theories. At the same time, the theory-driven approach has concentrated on cases where a strong theory could allow to derive rather precise predictions about the world that could then be tested. Some artificial intelligence work on theory revision have dealt with this situation (e.g. the Revolver system \citep{rose1989using}) but overall did not touch the problem of building a significant new conceptual domain.

In fact, the relatively few relevant works in machine learning (under the name of constructive induction, theory revision or inductive logic programming) have so far shared a common and mostly tacit assumption in that when learning a conceptual domain, the existing ontology of concepts was supposed to be correct, even if not always operationally efficient. The problem was thus seen as the one of  learning new concepts \textit{besides} existing ones, for instance by learning new concepts or predicates within the existing ontology in order to make it more efficient to use or more easy to understand. In this respect, the problem of learning a new conceptual domain was not really touched upon. In contrast, we see, in science and education, that one vital problem is to learn new concepts and new ontologies, at once articulated with past ones, but also in competition with them. For instance, learning the concept of ``particle'' in quantum physics implies keeping some of the aspects of the concept of ``particle'' in classical mechanics while erasing or deeply modifying some others. This is not to say that the `old' concept is erroneous or obsolete and must be discarded, but that its range of application must be circumscribed, and that it has to be superseded in some contexts and within some conceptual frameworks. 

It is rare, if it ever happens, that a new theory arises from scratch. Rather, it develops within an \textit{ecology} of other conceptual domains, more or less mature, and more or less operational and activated in the current context. There is therefore a complex interplay of articulation, facilitation, competition and hindrance taking place between various conceptual domains activated at the same time, and, focusing on the learning of a new domain, there are many places and many stages for transfers from one conceptual domain to the one in gestation (Figure \ref{fig:fig-1}). 

This paper reports on a multidisciplinary head-on approach to the problem of learning new ways to interpret the world by relying on (and relating to) old ones. By studying how high school students address problems in conceptual domains that are new to them, we were led to analyze mechanisms that seemed to be at play in their segmenting the world, and constructing models of the situation, as well as the (re)conceptualization efforts that —sometimes— followed. In this paper, we focus on a reasoning mechanism, that we hypothesize, does explain part of the students behavior. We call it \textit{tunnel effect} for reasons that will be clarified later on. Like analogy, this mechanism allows the transfer of knowledge from one conceptual domain to another one. Unlike analogy however, it does so without having to resort to two situations or cases, but only considers the one at hand, and it does not necessitate to specify beforehand a hierarchy of representation primitives in both domains (one being mostly unknown), nor to define how similarity between the two represented cases must be computed. In fact, it appears so natural that its scope covers a wide range of situations from metaphorical thinking to scientific discovery (See for other descriptions of historical cases of scientific discovery \citep{holton1988thematic,nersessian1992scientists,thagard1992conceptual}).   

\begin{figure}
  \centering
  \includegraphics[width=0.9\linewidth]{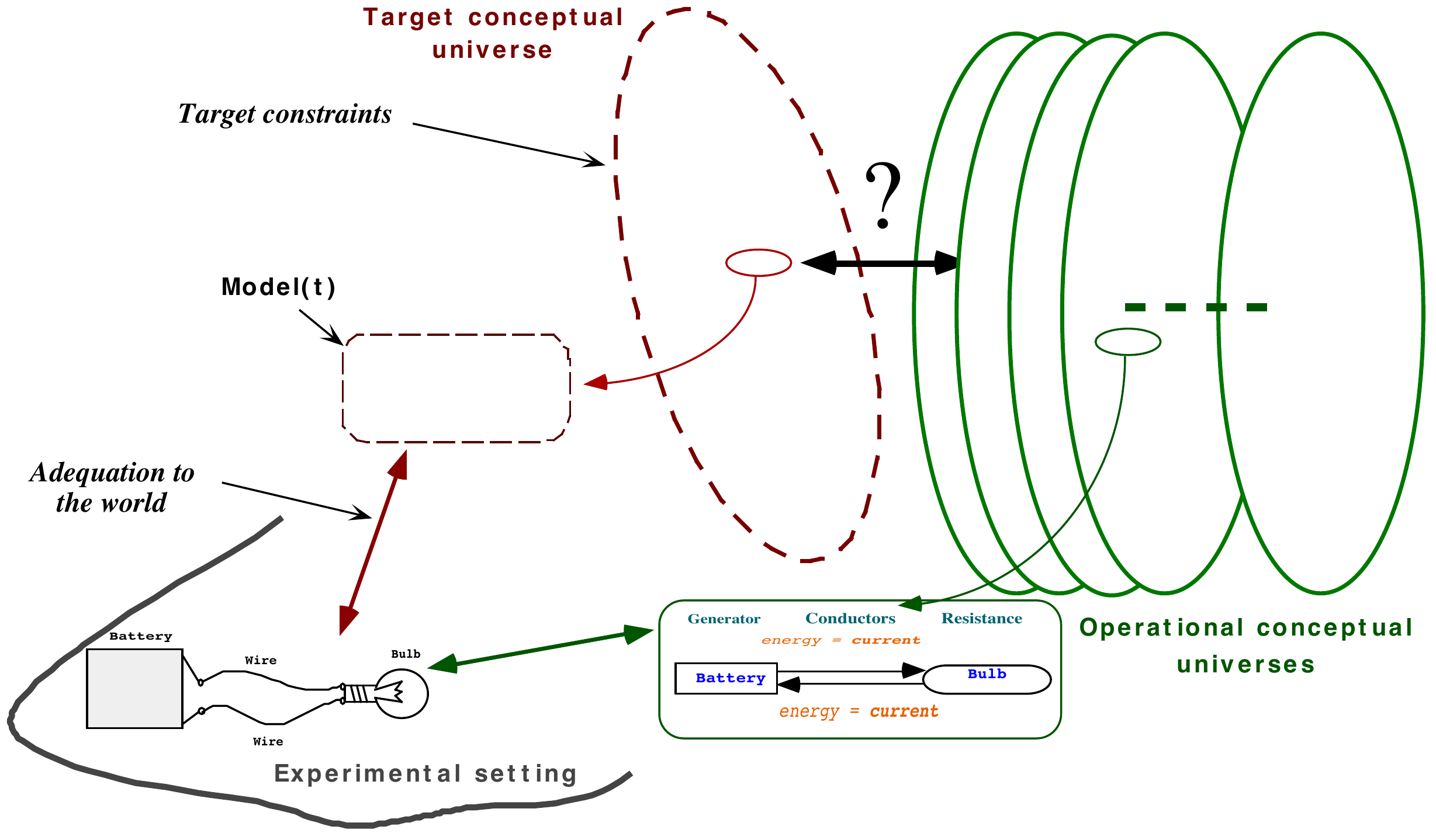}
  \caption{\textit{A view on the problem of learning a new conceptual domain}. A set of phenomena (e.g. a experimental setting) call for interpretation and explanation (some model) within a new target conceptual  universe. However, before reaching maturity and efficiency, the new conceptual domain cannot seamlessly and forcefully impose its own interpretation. Instead, there is an interplay with more operational conceptual domains that suggest their own viewpoint on the phenomena, for instance, here, the electricity domain with its circular model. In this paper, we study how learning a new domain occurs within the activity of an existing 'ecology' of other conceptual domains.}  
  \label{fig:fig-1}
\end{figure}

In the following, we first describe, in Section \ref{sec-illustration-tunnel-effect}, a simple but telltale experiment in physics teaching. Section \ref{sec-inferencing-mechanism} portrays how the mechanism of tunnel effect could account for it. A key ingredient of this effect lies in the hypothesis of the existence of a notional level in knowledge. This is analyzed in Section \ref{sec-notional-level}. How learning can occur as a result of tunnel effects is discussed in Section \ref{sec-learning}, while section \ref{sec-analogy-vs-tunnel-effect} contrasts tunnel effect with analogical reasoning with respect to the transfer of knowledge from one conceptual domain to another. Section \ref{sec-conclusion} concludes by underlying key ideas and perspectives.

\section{Illustration of the ``tunnel effect'' in physics teaching}
\label{sec-illustration-tunnel-effect}

Learning a new conceptual domain involves the interrelated processes of discovering potential target entities and mastering them in articulation with existing interpretation universes. This is the task faced by the scientist. In contrast, at school, most of the target conceptual domain is defined beforehand rather than having to be discovered anew, and it remains to provide it with meaning and potency. In the framework of studying the learning of a new conceptual domain, concentrating in a first stage on the latter task allows a better control of the learning assignment, making possible to manipulate the richness of the target domain, and its relationships with other, previously learned, conceptual domains.

\subsection{Design of the experiments}

In order to study learning of new conceptual domains, we set up interpretation tasks in terms of a ``new theory''. The idea was to force natural cognitive agents to learn a new way to interpret the world, and to study how they tend to do it. More specifically, we performed experiments in physics teaching, and more precisely teaching a qualitative account of the physics of energy taught in high school classes around the age 16-17. The task involved small experimental settings that the students could experiment with, like simple electrical circuits with masses and motors and so on, that were to be interpreted in terms of energy transfers and transformations along an ``energy chain'' starting and ending with an energy reservoir. The students worked in pairs\footnote{In fact the students are given successively three tasks, only the first task is discussed in the paper. In the first task the experimental material is made up of a bulb, two wires, a battery. In the second task the experiment consists of an object hanging on a string which is completely rolled round the axle of a motor (working as a generator). A bulb is connected to the terminals of the motor. When the object is falling, the bulb shines. In the third task the experiments consists in a battery connected to an electrical motor. An object is hanging from a string, attached to the axle of the motor, which is completely unrolled at the beginning. A correct solution is given to the students after the first task.}. This experiment has been done in several classes and in Andrée Tiberghien's laboratory. We video-recorded several pairs of students and entirely transcribed their verbal productions. For this paper, 7 pairs were deeply analyzed. (For complete details see \citep{megalakaki1995corpus,megalakaki1995experience}).

\begin{figure}
  \centering
  \includegraphics[width=0.95\linewidth]{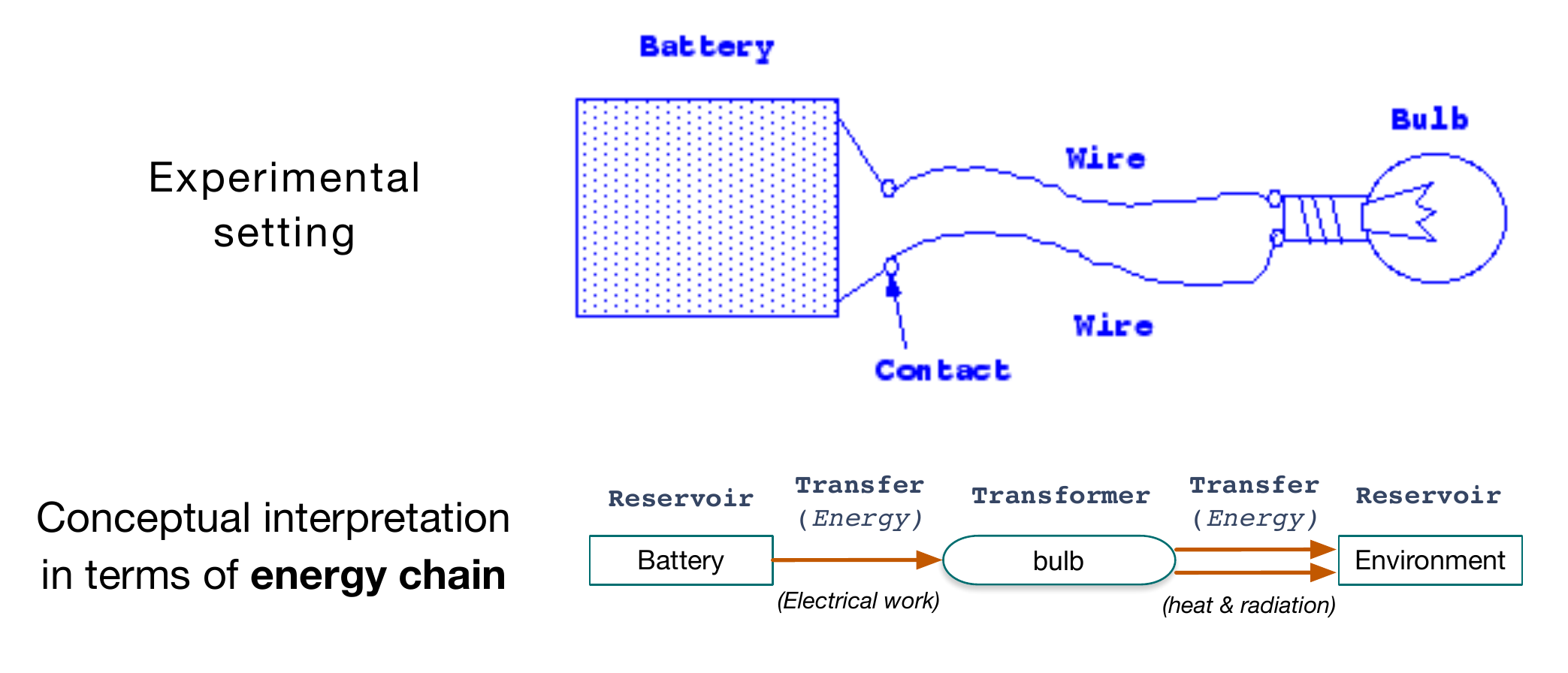}
  \caption{(Above): one experimental setting involving a battery connected to a luminous bulb through two wires. Students were to produce an interpretation of this setting in terms of a chain of energy transfers and transformations starting and ending with an energy reservoir. (Below): a correct interpretation, called target interpretation.}  
  \label{fig:fig-2}
\end{figure}

On one hand, it is important to notice that the interpretation task was not trivial, even in the simplest of the experimental settings shown in Figure \ref{fig:fig-2}. For instance, there were two wires  from the battery to the bulb which satisfied the closed electrical circuit condition, but only one counterpart, standing for the transfer of energy under the form of electrical work, in the target interpretation. Likewise, the students had to discover the environment entity while there was no concrete, tangible, counterpart in the experimental setting.

\begin{figure}
  \centering
  \includegraphics[width=0.9\linewidth]{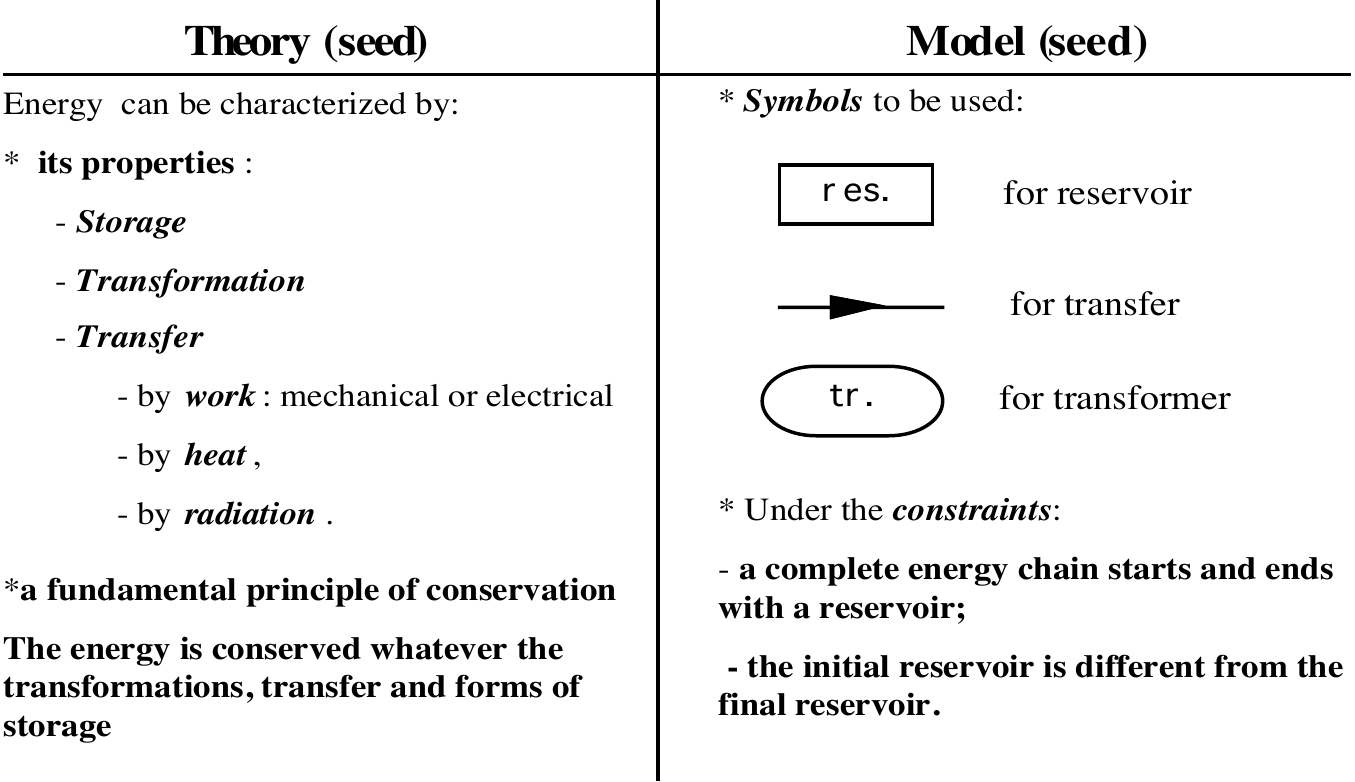}
  \caption{A simplified version of the seed for the target conceptual domain given to the students. The left part presents the conceptual definitions for the target domain. The right part provides the symbols with which to express the model and the syntactic rules that should be satisfied.}  
  \label{fig:fig-3}
\end{figure}

On the other hand, the task facing the students was easier than the one facing the scientists in that they did not have to ``invent'' the concepts necessary for the task. They were indeed provided beforehand with a declarative account of the target conceptual domain along with a lexicon of the authorized terms and icons that were to be used in their models of the situation (see Figure \ref{fig:fig-3}). This is one way we were able to control the knowledge brought to bear by the students. In particular the students were asked to use primitives like \textit{reservoir}, \textit{transformer}, and \textit{transfer}, for which they already had preconceptions (and how to do otherwise, except, may be, by some lengthy and convoluted paraphrase?). These preconceptions helped them to understand the seed theory, but of course they could also hinder later proper conceptualization.
The seed target domain also defined integrity rules that specified valid models, as, for instance, the ``a complete energy chain starts and ends with a reservoir'' rule.

Together, the \textit{lexical entities} used in the definition of the seed conceptual domain and the \textit{integrity rules} constitute the \textit{target constraints} for this particular task. 

\subsection{How the students can solve this problem}

\begin{figure}
  \centering
  \includegraphics[width=0.8\linewidth]{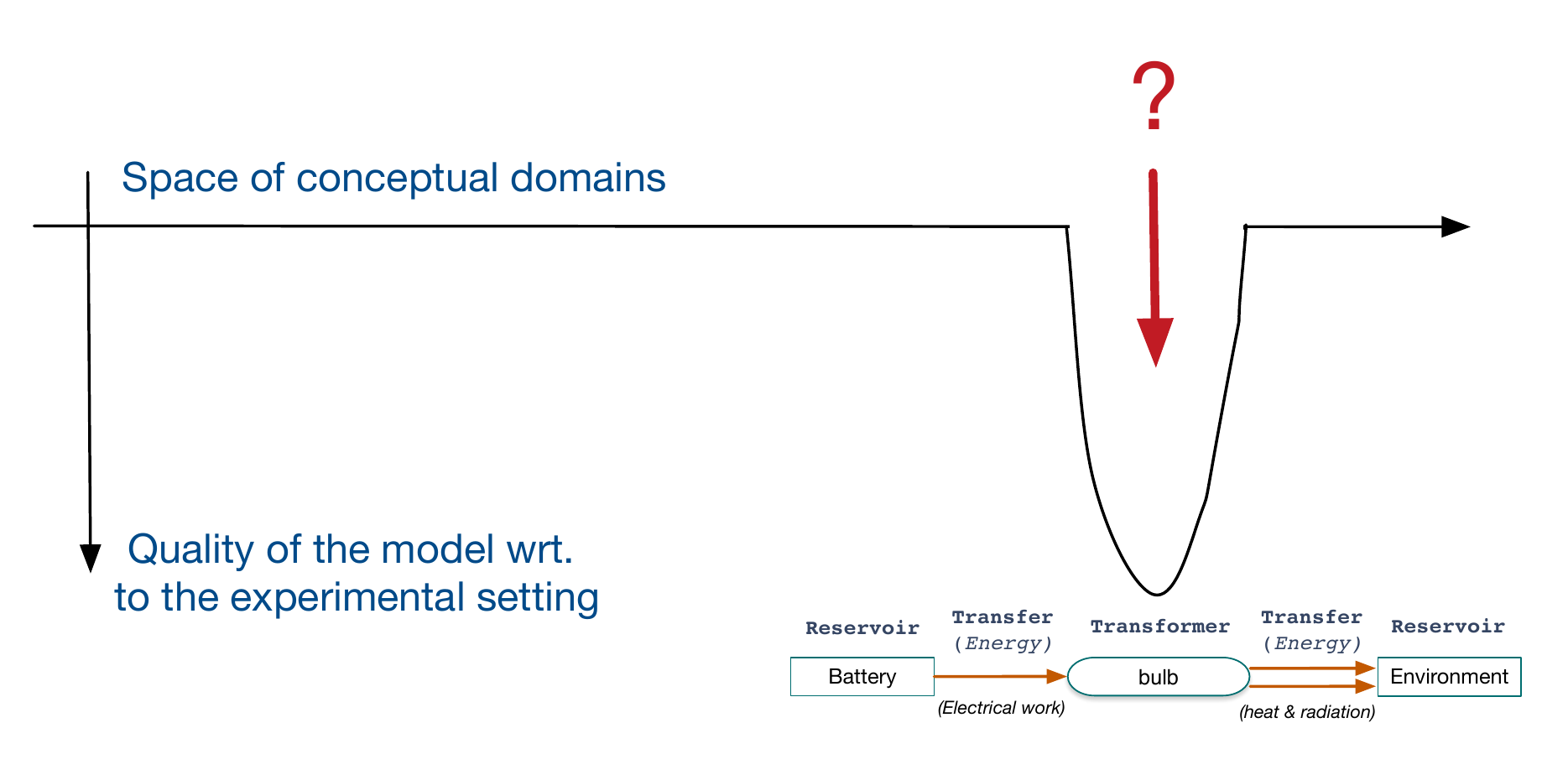}
  \caption{The horizontal axis stands for the space of potential models and the vertical axis stands for the quality of the model with regards to the world. The problem facing a newcomer to the target conceptual domain is to find the narrow well corresponding to the correct solution.}  
  \label{fig:fig-4}
\end{figure}

Now, we invite the reader to try for a few minutes to think of a program that could solve the problem above and others of the same type. Beware that the original description of the experimental setting is in itself a tricky problem. Some students for instance paid attention to details not shown here, like the electrical switch, the fingers, the eyes. Almost none however ``perceived" the environment as an entity. While all of them treated the two wires as two distinct entities, most did not single out the filament inside the electrical bulb. All in all, even in this extremely simplified setting, the perception and interpretation of the experiment involves an incredibly large collection of choices, both local and low level and global and strategic. Of course, when the target conceptual domain is well-mastered, as is usually the case for physics teachers, the interpretation task seems so easy that it is done seamlessly and almost unconsciously. It is then obvious that the ``correct and unique interpretation'' of the experimental setting is the one of Figure \ref{fig:fig-2}. But for a program to solve this interpretation task, how many choices to face, how much knowledge to have in order to make them efficiently! Figure \ref{fig:fig-4} schematizes this by showing a very narrow well in the landscape corresponding to all potential interpretations. This well indicates where one interpretation satisfies all of the target constraints. For a newcomer, finding it is like finding a needle in a bundle of straw. Is there any way that this can be made otherwise? Is there any way to help solve problems in an as yet ill-mastered domain?

One fact that emerged from our study was that out of 7 pairs of students, 6 produced the intermediate model of Figure \ref{fig:fig-5} (b) for the battery-bulb setting. They then departed from it to try to find alternatives, better suited models, meantime laboring over concepts like energy, transfers, and so on. This, in fact, did not strike us as worth of interest at first, so much it appeared to be expected. This intermediate model was after all none other than the classical circular electrical interpretation of the setting. Yet, upon reexamination, we were intrigued by the fact that this model, which acted as a powerful attractor, seemed also pivotal to enable further conceptual elaboration. Did the analysis of the why and how of this particular behavior could lead to a better understanding of the processes at play in the learning of new conceptual domains? The rest of the paper is an answer to this.

\section{The tunnel effect as an inferencing mechanism}
\label{sec-inferencing-mechanism}

\subsection{Analysis of the experiment}
\label{sec-analysis-of-the-experiment}

\begin{figure}
  \centering
  \includegraphics[width=0.96\linewidth]{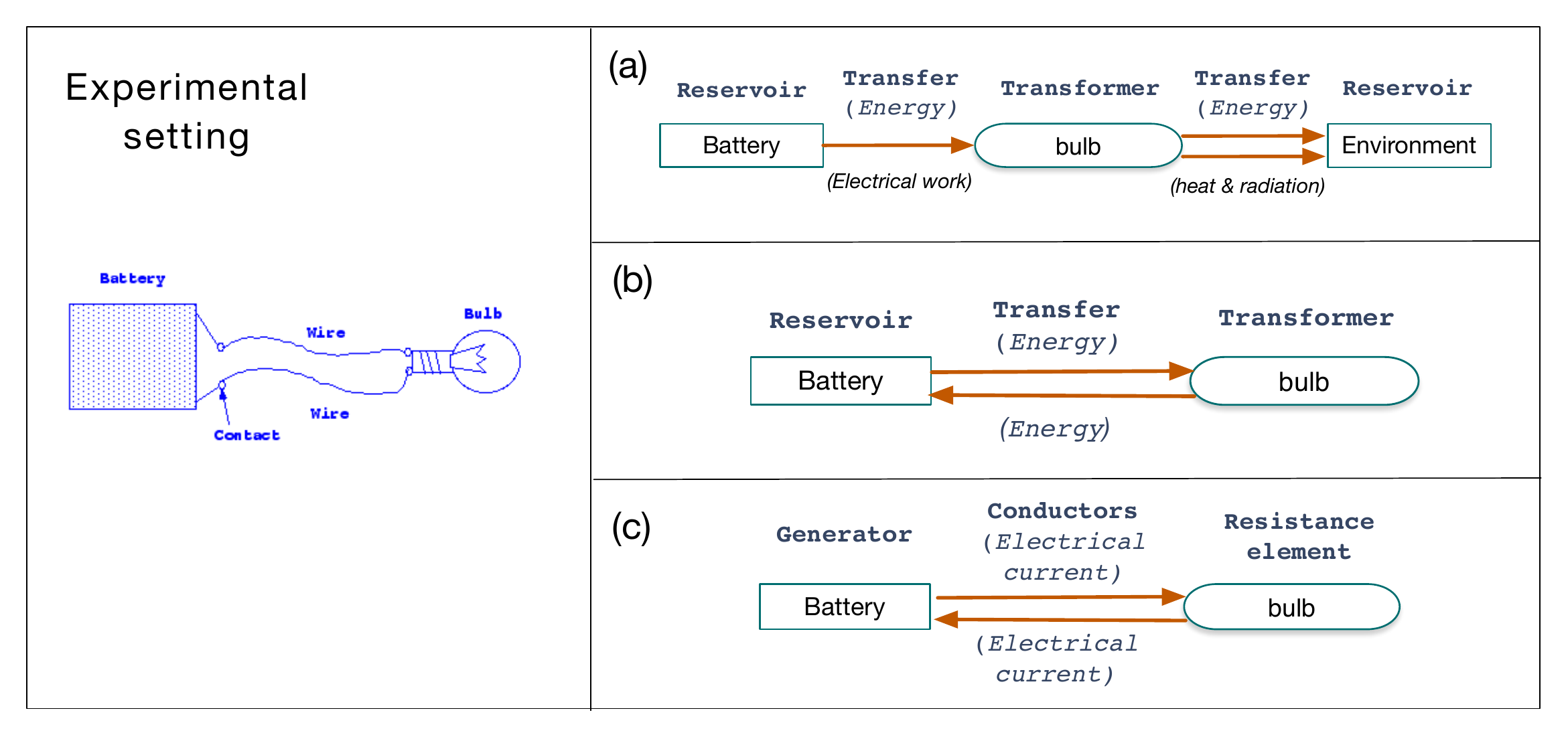} 
  \caption{Three interpretations of the experimental setting of the left column.}  
  \label{fig:fig-5}
\end{figure}

In the experiment described in Section \ref{sec-illustration-tunnel-effect}, had the students been experts in the domain of energy chains, they should have produced the model of Figure \ref{fig:fig-5} (a). Instead, the vast majority produced at some point the model of Figure \ref{fig:fig-5} (b), which clearly looks like model (c) which corresponds to an interpretation of the experimental setting in terms of electricity. However, they were not committed to an electrical interpretation, but were genuinely seeking an interpretation that would satisfy the constraints that were provided to them, that is an interpretation in terms of energy chains (cf. Figure \ref{fig:fig-3}). How to explain this discrepancy?

%

\begin{figure}
  \centering
  \includegraphics[width=0.3\linewidth]{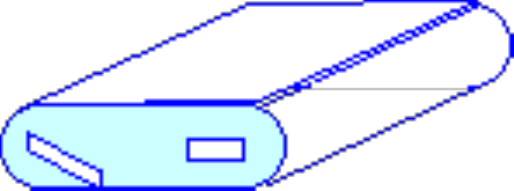}
  \caption{A shape that we almost necessarily designate by the category ``battery".}  
  \label{fig:fig-6}
\end{figure}

First of all, it is essential to realize that there is no such thing as an a priori unique, correct and complete representation of the reality. Any description of the world does depend on the subject's current understanding. However, there are descriptions that are very active in some situations and shared by a large set of individuals from the same cultural context. Thus, in our occidental culture of the turn of the third millennium, when someone sees some rather specific shape like in Figure \ref{fig:fig-6}, he/she cannot escape to see a battery, even if on an exploratory mission on Mars. 

Hence, depending on the context, there are categories, properties, relations and so on that literally impose themselves in the foreground. They are accordingly the ones that are used when communicating about the situation. Consequently, and quite naturally, the students in our experiments use words like ``the battery'', ``the wires'', ``the lamp'' to point out things in the world. Not only do they use these words to communicate, they also think about the situation using the associated entities. For instance, suppose you are on an first exploratory mission to Mars, and you see a few steps ahead of you a shape like in Figure \ref{fig:fig-6}. Not only will you describe it as a battery, but you would seem literally out of your mind if you did not also jump out of surprise and start considering all the implications, including the possibility of the existence of some extraterrestrial life forms knowing of electricity and having set foot on Mars. Referring to some set of perceptions using some words indeed goes far further than merely uttering a designation, it brings with it a whole lot of expectations, constraints on the world, and generally associations to other conceptual structures.

If we then look at the process by which the students built their interpretation of the experimental setting, they start by trying to recognize the matches between the outstanding categories they identify like ``the battery”, ``the wires'' and ``the lamp'' and the target concepts that have been given to them in the seed theory. They thus match without much trouble the battery with a \textit{reservoir}, and the lamp with a \textit{transformer} . Without entering into details here (see Sections \ref{sec-notional-level} and \ref{sec-learning} below), this is rather easy because, on one hand, batteries and reservoirs share a lot of common properties like being subject to be full or empty, or to play a causal role in many situations, and, on the other hand, lamps transforms electricity into light (and heat), and thus is an instance of a transformer.

Now, when they come to consider the wires, it is clear that in the context of this experiment –a physics course and a typical example of an electrical circuit–, the most operational and highly activated interpretation domain is the electricity domain as it has been previously learned in the physics classes. Consequently, and again without entering into details, the wires are matched with \texttt{Means\_for\_Energy\_Transfers}, and the electrical current is matched with \texttt{Transferred\_Energy}. This electrical counterpart to \texttt{Energy\_Transfers} allows to import the automatically inferred (within the electricity interpretation domain) directions of the electrical currents and to make them also the directions for the \texttt{Energy\_Transfers} . 

It is interesting to see that in this process, compelling interpretations of the world, related to batteries, lamps and electrical circuits have naturally been incorporated into the built model, they have also completely shaped it. Thus, in particular, the commanding electrical interpretation of the setting has led to the model (b) of Figure \ref{fig:fig-5} which is akin to the electrical interpretation of model (c). And in this way, lots of aspects and inferences from the activated interpretation domains that are foreign to the target one have nonetheless  been illegally imported  insofar as they provided means to fill up the missing roles of the target model. 

Thus the overall interpretation, even though it is based on foreign pieces of interpretation and is deeply of an electrical nature,  fits most of the syntactical constraints for the satisfaction criteria, and can be \textit{mistaken for} a valid energy chain interpretation of the setting. 

\begin{figure}
  \centering
  \includegraphics[width=0.86\linewidth]{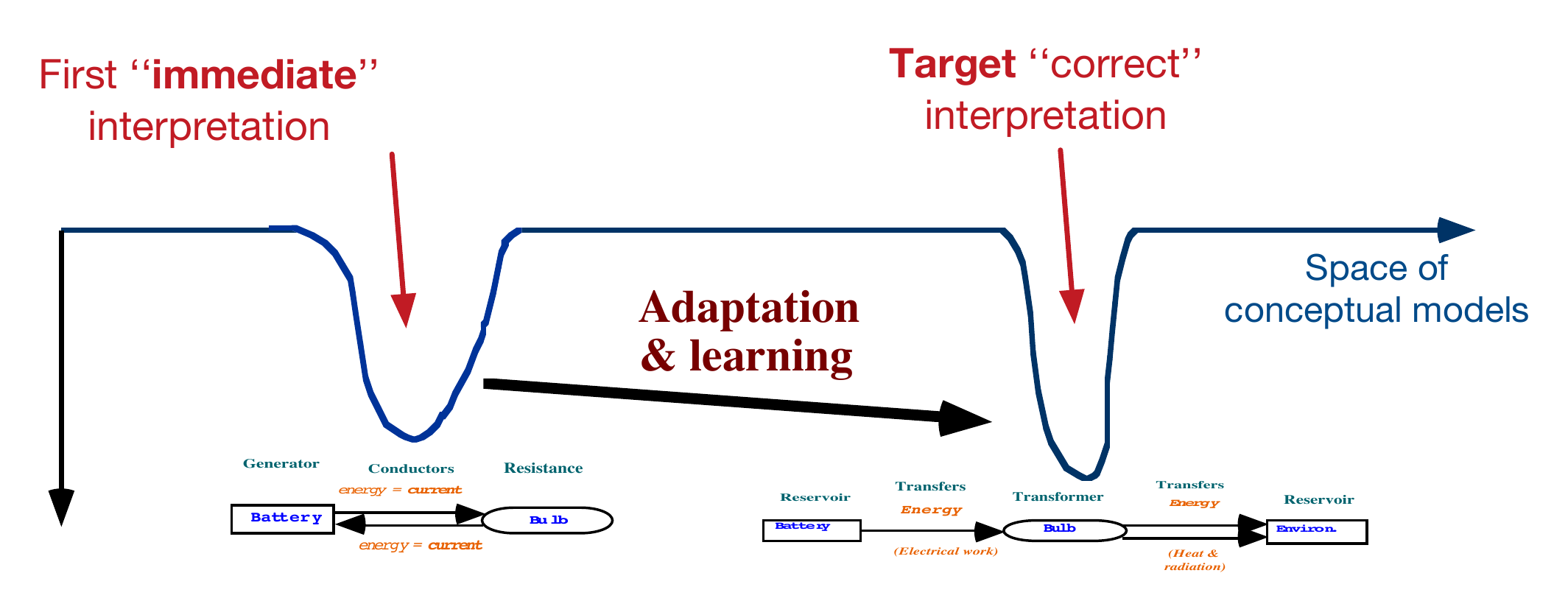}
  \caption{The cognitive tunnel effect in effect allows the learner to use its first ``immediate'' conceptual interpretation of the world in order to construct the ``correct'' model, thus bypassing a lengthy and difficult search in the vast space of conceptual models.}  
  \label{fig:figure-tunnel-effect-2}
\end{figure}

But if the interpretation activity stopped there, that would not lead to any further ``discovery'' by the cognitive agents, nor to the reconceptualization that is part of the learning of a new conceptual domain. There would simply be an electrical interpretation of the phenomenon disguised as an energy chain one. What is interesting is that, except in one case, all the students then embarked in further predictive activity based on the model they just came up with. They thus reinterpreted the model stemming from electrical considerations within the energy interpretation domain! That is they suspended the underlying raison d'\^etre of the model to interpret it within a new conceptual domain. This seemingly instantaneous autonomisation of the model and reinterpretation within a new conceptual domain, is reminiscent of the tunnel effect in quantum physics whereby a particle can occasionally tunnel through barriers, or, more precisely, escape a potential well to enter another one without having enough energy to overcome the potential barrier between them. In our case, the passage from one interpretation domain to another one should go along with a dismantling of the first interpretation/model before reconstructing a new one in the new domain. Instead, the very same model becomes autonomous from its former source domain and is reinterpreted as such in a new one as if a tunnel had been drilled between the two domains allowing one to go from one to the other unnoticed and inconspicuously (see Figure \ref{fig:figure-tunnel-effect-2}). Before analyzing how this is possible, it is interesting to see what happens next. 

Some students, ``reading'' the consequences of their model, predicted that energy will come back to the battery, which they know, from their prior background knowledge, is not possible. These students are thus in front of a violation of the adequacy to the physical world criterion. Others (a minority) realize that their model does not satisfy the integrity rule of the seed theory according to which the initial and final reservoirs should be different. In any case, an intense reexamination of the model, its meaning and its justification takes place, leading to a better mastering of the target domain (\citep{cauzinille1997co} provides an analysis of the ``repair mechanisms'' used by students to adapt their model). 

\subsection{Analysis of the tunnel effect mechanism}

In this Section, we present five key ingredients that concur to the tunnel effect inferencing mechanism. Our description is oriented towards the specification of a computational model.

\medskip
\noindent
\textbf{1. The target conceptual domain is specified by target constraints}
\smallskip

It is important to realize that, in scientific discovery as well as in education, even though the target domain may still be unknown, there usually exists some a priori constraints that delimit it. These can be found at two levels.

\medskip
\textit{First}, a level concerned with \textbf{meta-constraints}. In particular:

\begin{itemize}
   \item \textit{Syntactical constraints}. When mature, a conceptual domain can function entirely as a closed system with entities entertaining relationships with other entities of the same domain and defined only within this domain. These relationships may then be characterized by syntactical constraints that define rules for well-formed formulas and for valid derivations. For instance, in the simplest of cases, these may only consist of dimension equations that have to be satisfied. The right side of the seed theory for the energy domain is another example of such syntactical constraints.
   
   \item \textit{General a priori commitments}. It is generally the case that the founders of a new scientific domain are guided by a priori deep beliefs regarding the world (See for instance \citep{holton1988thematic}  for an interesting discussion on some themata that steer scientific investigation. \citep{harman1982energy} provides also analyses on the underlying philosophical commitments of searchers during the 19th century). These preconceptions may have some repercussions on syntactical constraints. For instance, while developing his theory of electromagnetism, Maxwell was deeply influenced by the concept of continuous action in a medium as opposed to the Newtonian idea of action at a distance that was then the guiding concept in France. Accordingly, Maxwell was naturally looking for a theory expressed using differential equations. 
\end{itemize}

\textit{Second}, a level corresponding to the \textbf{adequacy to the world}.  To be viable, a conceptual domain must possess a good adequacy to the world, that is be able to allow coherent and sufficiently complete descriptions of the world and make predictions that are reasonably confirmed. 

Two consequences follow from the existence of these meta-constraints. Indeed, on the one hand, any model or theory must obey them in order to be valid. On the other hand, any model or theory satisfying, at least at first sight, these meta-constraints may appear to be valid. This second part is critical for the tunnel effect.

\medskip
\noindent
\textbf{2. Interpretative systems imply active entities}
\smallskip

The central task we study in this research is the one of interpreting the physical world according to some conceptual domain given data that can be incomplete and imperfect in various ways. This type of interpretative tasks is far from uncommon and has been one focus of AI research particularly in the field of vision and of natural language processing. Most works have ended up by underlining the role of active structural networks where knowledge is composed of semantic entities organized along relevant semantic links. Each entity, is itself a small organization with pointers to other potential entities and inferencing mechanisms that allow to identify these according to the context. In the language of \citep{minsky1975framework,rumelhart1976accretion}, and \citep{schank1983dynamic}, these entities are called \textit{schemata}, the pointers \textit{slots} and the associated inferencing mechanisms \textit{demons}.
 
We believe this view to be pertinent as well in the context of the interpretation of the physical world. This is why we envision conceptual domains as composed of schemata, actively engaged in the comprehension of arriving information and guiding the execution of processing operations. 
 
\textit{Generic concepts are represented by schemata}. These contain variables: references to general classes of concepts that can actually be substituted for the variables in determining the implications of the schema for any particular situation. Inference mechanisms, often local to the schemata, are responsible for this. Particular information on the current situation being interpreted is encoded within the memory system when constants --specific values or specific concepts-- are substituted for the variables of a general schema. This is for instance what happens when \textit{battery} is substituted for the \texttt{initial\_reservoir} variable of the general schema corresponding to \texttt{energy\_chain}.
 
Many of the variables in a schema can have default values associated with them. Their attached demons can also determine potential substitutions. In that way, the whole system is able to encode the situation at hand, to interpret it, and to make predictions about it. This is exactly what we expect from the system.

We will now see what role these schema may play in the tunnel effect mechanism.

\medskip
\noindent
\textbf{3. Candidate entities within the target domain may become associated with known entities (Janus entitites)}
\smallskip

During the first attempts to account for sets of phenomena in terms of a conceptual domain in construction, it is natural and usual that ill-known target entities be thought of in terms of known entities from other, more operational, conceptual domains. 

Hence, it is remarkable that students incessantly use one term for another (for instance ``energy'' and ``(electrical) current'') as if they were interchangeable. This certainly testifies of some underlying confusion regarding these notions. The very same type of confusion seems to be at play in the origin of many scientific domains. For instance, the concept of \textit{heat} was painfully distinguished from the concept of \textit{caloric}, itself associated with hydraulic connotations. Likewise, the concept of \textit{speed} was for a long time intimately associated with the concept of the \textit{force} causing the movement, thus making difficult, if not impossible, to consider changes of referential (see for instance \citep{viennot1996raisonner} [Viennot,1996] for this and other examples). Any new conceptual domain is learnt in interaction with existing interpretative domains. New concepts can be defined in terms of known concepts, like the concept of mass in General Relativity that Einstein derived from a combination of inertial mass and gravitational mass. They also may be mistaken with concepts from other conceptual domains within a single undifferentiated entity, as when the students associate \textit{energy transfers} with \textit{electrical current}, or when Sadi Carnot, in the 1820s, adopted the caloric interpretation of heat. We call these undifferentiated entities \textit{Janus entities}\footnote{Roman god represented with two opposite faces. In Roma, he is the guardian of the gates (\textit{januae})!!} for they are the two faces of a single functioning entity as we now see.

\medskip
\noindent
\textbf{4. Once associated, the entities may share their inference mechanisms}
\smallskip

We postulate that entities that are associated within a Janus entity may, if needed, share their inferencing mechanisms. For instance, once \texttt{Energy\_Transfer} and \texttt{Electrical\_Current } are associated within one Janus entity, for they are used equally by the students, when it becomes necessary to find the direction of \texttt{Energy\_Transfer}, this is the inferencing mechanism associated with the direction of \texttt{Electrical\_Current} that is triggered. Hence, if this mechanism determines that the  directions are $\rightarrow$ and $\leftarrow$  on account of the $+$ and $-$ terminals of the battery, then these directions are assigned for the \texttt{Energy\_Transfer}, even though the underlying reasons are foreign to the energy domain. These inference procedures are not thought upon and pondered while introduced and used in the new conceptual domain, but, on the contrary, they are in a way \textit{smuggled in without further immediate checking}. 

\begin{figure}
  \centering
  \includegraphics[width=0.55\linewidth]{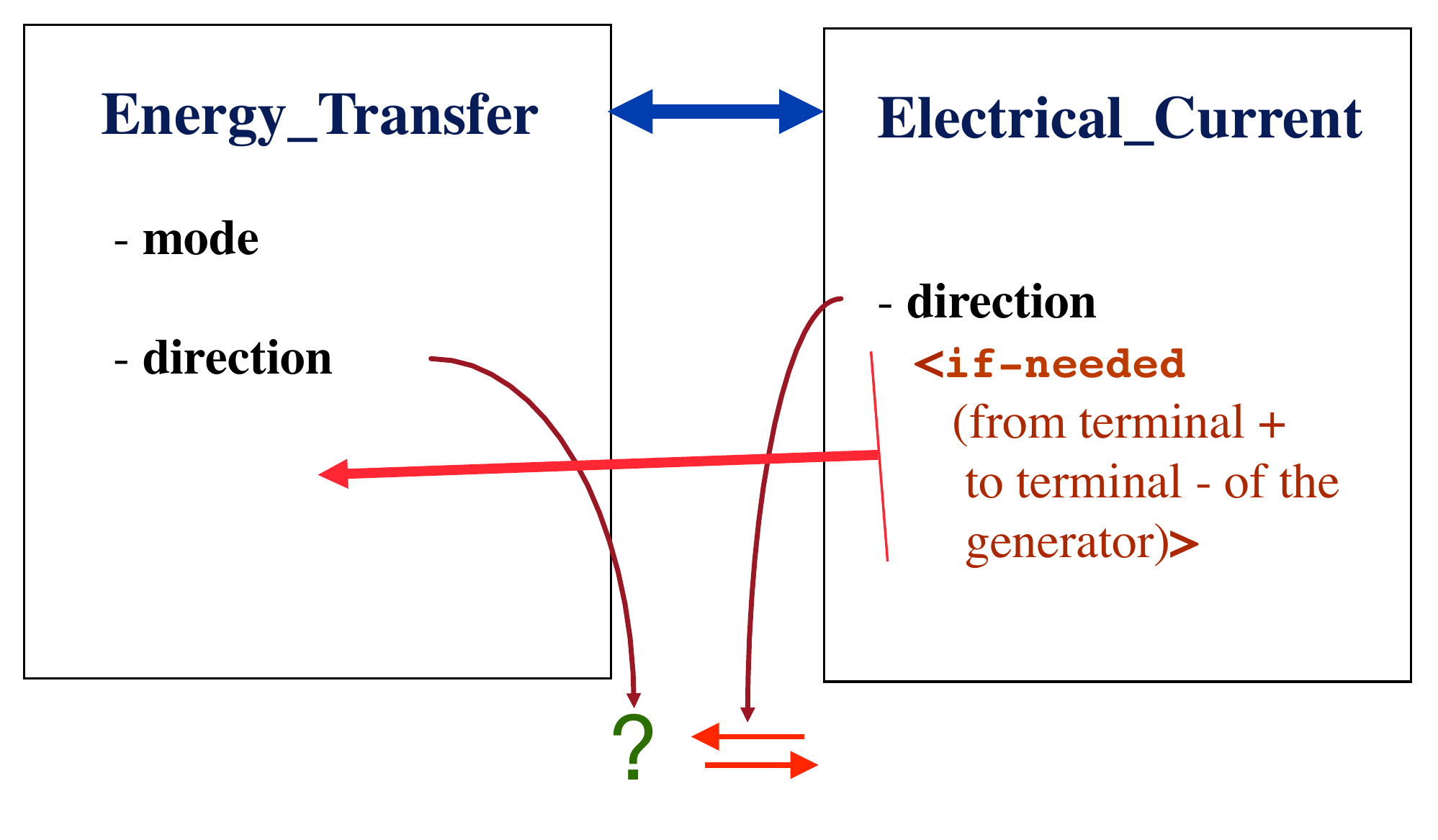}
  \caption{When two entities, shown here as conceptual schemata with attached slots or aspects and inferencing mechanisms, are associated within a Janus entity, they can share aspects and methods. For instance here, when there is a need for the determination of the direction aspect of \texttt{Energy\_Transfer}, the method from the corresponding aspect of \texttt{Electrical\_Current} can be used in place of the missing one.}  
  \label{fig:fig-1}
\end{figure}

It is important to realize that this phenomenon, which is central in what we call the tunnel effect in cognition, is ordinary. It happened when Carnot was equating the \textit{caloric} with \textit{heat}, thereby introducing --smuggling in-- its conservative property and thence its cyclic character. This model, formalized by Clapeyron and later thought upon prominently by Thomson and Clausius, is completely unexplainable devoid of this underlying commitment, that is unthinkable in a world where, thanks to Joule, heat and work are two forms of energy, interchangeable in part. It happened to Maxwell when he equated the ether (incompressible fluid) with a model for electromagnetic interactions, smuggling in the seeds for the difficulties faced in physics until Einstein's special relativity theory got rid of them (and of most of the smuggled in properties of ether). It happens all the time, and it happens unconsciously. This smuggling might turn out to be genial when it brings with it unexpected solutions to outstanding problems. There is no reason why it might not also hinder further solution.

To sum up, each time entities from two different interpretation domains are matched or associated within one Janus entity, they can bring with them:

\begin{itemize}
   \item [(i)] \textit{Inferencing mechanisms} corresponding to common aspects of the associated entities. For instance, the demon attached to the direction aspect of \texttt{Electrical\_Current} can be used in place of the --still missing-- corresponding demon for the direction aspect of \texttt{Energy\_Transfer}. 
  
   \item [(ii)	] \textit{New aspects}. For instance, in another task not detailed here, one student made the association between \texttt{Energy\_Reservoir} and \textit{weight}. This in turn allowed him to show how to ``fill up the weight" by rising it! This is one clear instance of a property --to be fillable-- originally absent from the weight concept, but brought over by the association with \texttt{Reservoir}.
\end{itemize}

\medskip
\noindent
\textbf{5. The reinterpretation of the built model entirely within the target domain may produce unforeseen and unpredictable consequences.}
\smallskip

The mechanism described above that is responsible for the sharing of aspects and inferencing procedures may lead to a true transfer of information from one conceptual domain to another when the interpretative model of the world built in part thanks to the former is then reinterpreted entirely  within the target domain. 

For instance, in the task described in Section \ref{sec-illustration-tunnel-effect}, the students built the circular model shown in Figure \ref{fig:fig-5}(b) using inferencing mechanisms from the electrical face of the \texttt{Electrical} \texttt{\_Current}/ \texttt{Energy\_Transfer} Janus entity. They then undertook to interpret this model in order to make predictions: for instance they predicted that the energy goes back to the battery, or better still that all of the light coming out from the bulb goes back entirely to the battery, which is in fact what the model expresses. There is therefore no doubt that this interpretative activity took place entirely within the target domain, that is the energy domain. 

Besides, it must be realized that the circular nature of the model of Figure \ref{fig:fig-5}(b) could not be obtained within the energy conceptual domain. There simply would be no reason for it. It had to be built using some information from other domains. We have already stressed the same point regarding the Carnot's model in thermodynamics, which, stemming from foreign preconceptions (heat viewed as an imponderable and conservative substance), was reinterpreted within the new thermodynamics yielding the concept of entropy, a strange state function defined with respect to the ideal cycle of Carnot. \citep{harman1982energy,locqueneux1996prehistoire} and \citep{stengers1996cosmopolitiques}, among others, have excellently shown how the circular model of Carnot would have been impossible to conceive of in the new thermodynamics of Joule. This is a clear example of a transfer from one conceptual domain to another. The genius of Carnot lies in part in having proposed a model, that even though was founded on erroneous postulates (the caloric hypothesis), could be reinterpreted fruitfully in the new domain. This transfer of a model from one conceptual domain to another was possible because the meta-constraints were compatible. In particular the syntactic ones (pressure, volume, temperature and so on regarded as relevant variables) were the same. And, at a deeper level, Carnot's model was a way to express and test the principle according to which the cause was conserved in the effect, cause and effect being either some matter (the caloric) or some force (the so-called living force, later called kinetic energy). 

\medskip
In summary, the tunnel effect transfer mechanism relies on the following elements:

\begin{enumerate}
   \item Each conceptual domain, be it in gestation, is specified by \textit{meta-constraints} that set in particular how a valid model should look like as a syntactical construct. 
   \item Conceptual domains are made of active conceptual entities that can be modeled as \textit{schemata} in a semantic network.
   \item Target conceptual entities can in a first stage be intimately associated (confused) with entities belonging to other conceptual domains. These associations are called \textit{Janus entities}.
   \item Janus entities allow each of the associated entity to \textit{borrow aspects and inferencing mechanisms} from their twin face when needed. This may help the building of a model of the situation to be interpreted.
   \item If and when the build model is \textit{reinterpreted} within the target conceptual domain, which is made possible if and only if the meta-constraints of the target domain are satisfied, then unexpected conclusions may ensue. In this case, it can be said that a transfer of information has occurred between some source domain(s) and the target domain. 
\end{enumerate}

\subsection{Why call it tunnel effect?}

\begin{figure}
  \centering
  \includegraphics[width=0.9\linewidth]{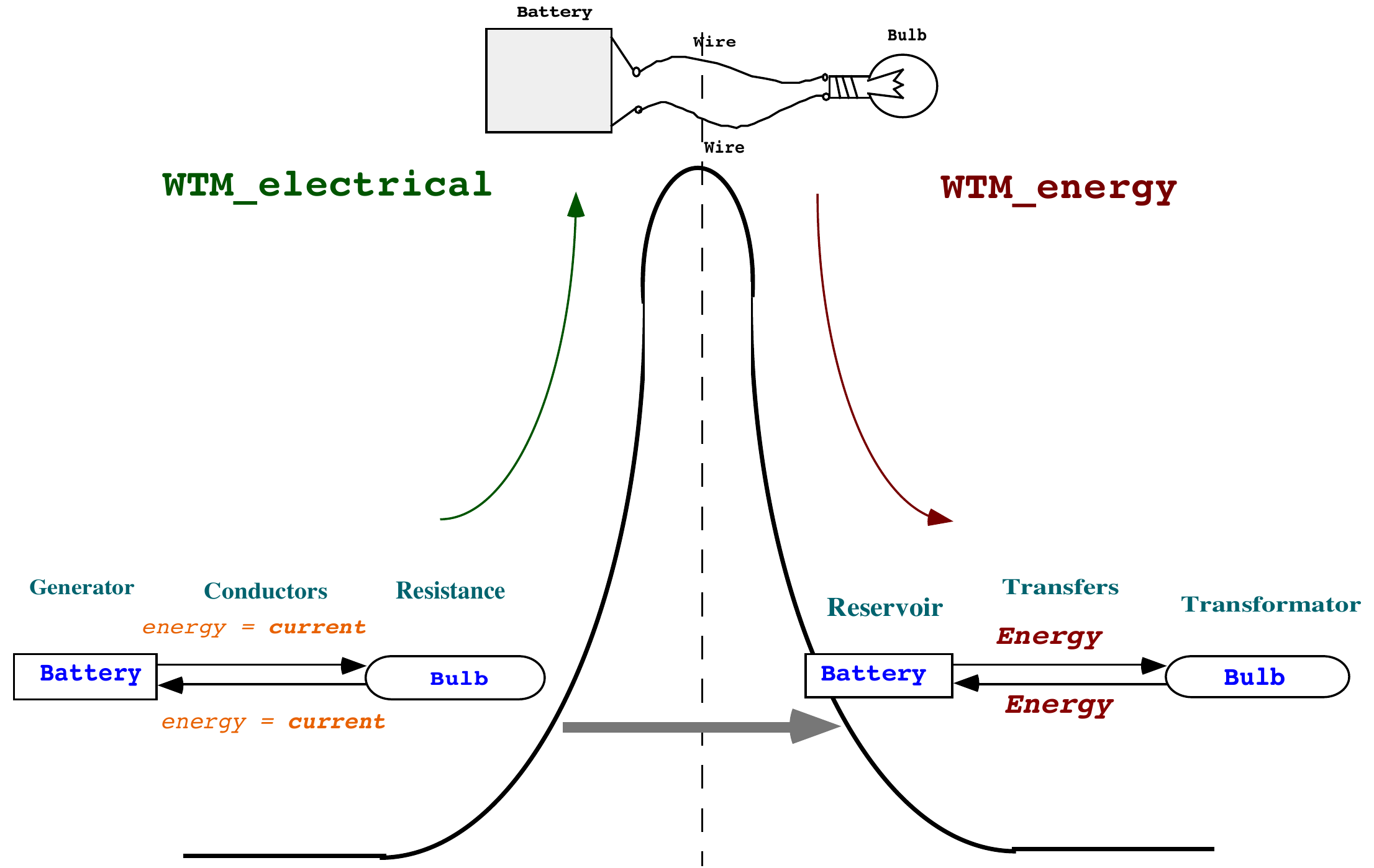}
  \caption{The way a situation should be reinterpreted when going from one conceptual domain to  another is to abandon the first interpretation, come back to the original situation and build a new interpretation in the new domain (as indicated by the curved arrows above the barrier). In the cognitive tunnel effect, as in the quantum tunnel effect, the interpretative model built in one conceptual domain may be directly imported from the original domain to the new one (gray arrow). In this way, it is possible that some information properly belonging to the original domain are smuggled into the new one without checking. This is all the more feasible that the metaconstraints corresponding to each conceptual domains are similar, thus making possible that the original interpretation be mistaken for a valid one into the new domain.}  
  \label{fig:fig-tunnel-effect}
\end{figure}

The cognitive mechanism described above has been named \textit{tunnel effect}  because of an analogy with a phenomenon in quantum physics whereby sometimes a particle may apparently jump over a barrier between two potential wells without being endowed with the necessary energy to do so (see Figure \ref{fig:fig-tunnel-effect}). We feel that, in a way, it is possible to envision each conceptual domain as a different potential well. A situation (for instance, a experimental setting) may be interpreted within one or another conceptual domain. But to reinterpret the situation in terms of a different conceptual domain should require to abandon the initial interpretation (expressed as a model in the initial domain), to go back to the situation itself and then to build a new interpretation in the new domain. The transfer mechanism we call tunnel effect acts as if there was no barrier between the wells. The original model built inside the first domain is thus \textit{taken and imported in the new domain disguised as if it was a licit model}. This is possible when the meta-constraints specifying the two domains are compatible. The thus transferred model might bring with it unchecked aspects and information that are foreign to the target domain. As we have seen, this in turn opens the possibility for unpredictable consequences when reinterpretation occurs in the new domain. 

After the description of the tunnel effect mechanism, several questions immediately arise, namely, where do Janus entities come from? How do different entities can become associated? How the tunnel effect can affect subsequent learning? Is tunnel effect associated with a special form of learning? 

In the following, we study these questions in turn

\section{The notional level hypothesis}
\label{sec-notional-level}

\subsection{The hypothesis}

Gentilhomme (1994), a linguist interested in the ``techno-dialects", has proposed that a distinction be made between \textit{notions} and \textit{concepts}. Following this line of investigation, we introduce the hypothesis that two levels of description co-operate while constructing a new conceptual domain. 
In the context of this study, the relevance of this idea is threefold: 

\begin{enumerate}
   \item It explains how conceptual entities, each properly belonging to closed conceptual systems, can be associated.

   \item It supplies a source for the origin of candidate target entities, that are otherwise difficult to account for.

   \item It provides a flexible basis for learning concepts: by specializing notions.

\end{enumerate}

The hypothesis put forth by Gentilhomme is that \textit{there is a difference between an informal, common sense use of language, and a rigid formal one used in scientific discourses}. He argued that a scientist, in his/her professional activity, is always playing back and forth between these two extremes, usually happily using the scientific concepts of his trade, with all their precision and attached apparatus of constraints and methods, but sometimes going back to the fluid common sense notions when dealing with ill-mastered domains. For instance, a physicist would naturally resort to the \textit{notion} of \textit{twisting} when trying to understand how the \textit{concept} of \texttt{torsion} could apply in a new domain. 

In this way, the notional core attached to each concept would \textbf{provide a bridge} for migrating from one conceptual domain to another. 

\begin{figure}
  \centering
  \includegraphics[width=0.8\linewidth]{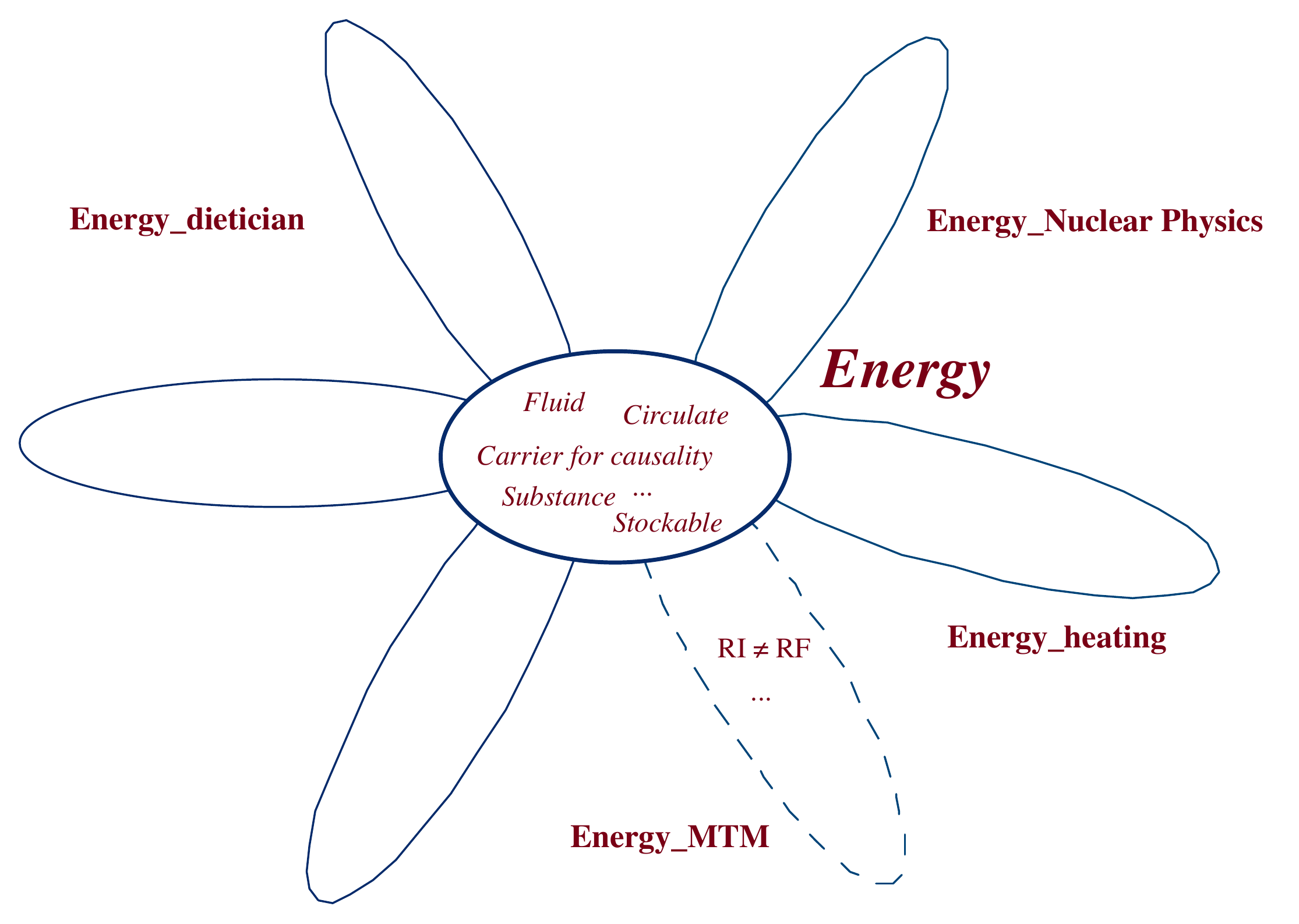}
  \caption{The notion and concepts associated with energy.}  
  \label{fig:fig-7}
\end{figure}

Figure \ref{fig:fig-7} tries to give a qualitative picture of this idea. For each linguistic entity, there would be a notional core made up of the common sense notions associated with it. For instance, the notion associated with ``energy" would include beliefs (represented as schemata in our model (see Section \ref{sec-learning-concepts-from-notions}) with energy as a role) such as: \textit{energy is a kind of fluid or substance that can circulate, be stored} (e.g. the expression ``I am full of energy this morning"), and is often linked with causality. Around this notional core, there would be several ``petals", each one corresponding to some specific conceptual domain resulting from a deliberative conceptual work. In the energy example, petals could be associated with the definition and properties of energy as elaborated in nuclear physics, or in dietetics, and so on.

\subsection{Criteria for characterizing notional and conceptual levels}

Notions and concepts can be distinguished from a linguistic point of view, considering the properties of the linguistic items themselves, and their behavior pattern within dialogs. Though everyday language and scientific language cannot easily be separated from each other, some criteria are still fit for use.

\medskip
\noindent
\textbf{(a) Shift for meaning}
\smallskip

The notional level requires a fixed meaning, shared by all the speakers. One must pay attention to this meaning in order to be understood, but figures of speech are still possible and quite frequent, such as metaphor, metonymy, play on words. On the contrary, the meaning of scientific terms, that belong to conceptual level, is more arbitrary, based on a conventional definitions ; but it requires absolute faithfulness to that definition.

The following extract of a dialog proceeding from the energy task, points out a metaphoric use of the word “energy”:

\vspace{-0.3cm}
\begin{center}
\line(1,0){250}
\end{center}
\noindent
\small
\begin{tabular}{ll}
F	& \texttt{Energy is moving, what would you say?}  \\
$[...]$ & \\
F	& \texttt{Energy just goes through…}  \\
$[...]$ & \\
F	& \texttt{Energy comes back from the bulb.} \\
$[...]$ & \\
F	& \makecell[tl]{\texttt{You see, energy leaves from there.} \\ \texttt{It goes “ttssouiii” through the wires, and gets there.}} \\
\end{tabular}
\begin{center}
\line(1,0){250}
\end{center}
\normalsize

\medskip
\noindent
\textbf{(b) Re-formulation, periphrasis, and synonymy}
\smallskip

The notional level allows the use of synonyms and various re-formulations. Things which are forbidden by the rules of scientific speech, since there cannot be any perfect equivalent to a given term. Therefore, any attempt to grasp the meaning of a given term by way of paraphrases can be seen as a symptom that the agent is staying at the notional level.

\smallskip
\noindent
For example, searching for an energy transfer, a student states:

\smallskip
\small
\begin{tabular}{ll}
F	& \texttt{The transport modes} {(instead of transfer modes)},\texttt{ it's not only heat ...} 
\end{tabular}
\smallskip
\normalsize

\smallskip
\noindent
And further:

\smallskip
\small
\begin{tabular}{ll}
F	& \makecell[tl]{\texttt{Yes, but inside the battery, is it hot, inside the battery?} \\ (using hot instead of heat)} \\

\end{tabular}
\normalsize
\smallskip

\smallskip
\noindent
Searching for an energy reservoir, the same student says:

\smallskip
\small
\begin{tabular}{ll}
F	& \texttt{to stock… the reservoir, it’s where the reserve is, I presume ...} \\
\end{tabular}
\medskip

\normalsize
Similarly, the use of antonyms can help differentiate the two levels. As a matter of fact, the opposite of “acid” is not the same at the conceptual level in chemistry (“basic” or “alkaline”), and at the notional level where it could be “sweet” as well as “gentle” or “soft”.

\medskip
\noindent
\textbf{(c) Translation}
\smallskip

A term related to a scientific concept has to be translatable into another language in a one-to-one way. Such is not the case for a word used at a notional level.

\subsection{How entities from different domains can be associated: Janus entities}
\label{sec-janus-entities}

There is plenty of evidence that human subjects smoothly associate entities belonging to different and in some sense incommensurate domains. Considering only the dialogs registered during the energy chain task described in Section \ref{sec-illustration-tunnel-effect}, we get instances like the following ones (see  \citep{megalakaki1995corpus,megalakaki1995experience}) where different entities are matched or used as if they were equivalent.

\vspace{-0.5cm}
\begin{center}
\line(1,0){250}
\end{center}
\noindent
\small
\vspace{-0.2cm}
\begin{tabular}{ll}
Martin (27) &:	\texttt{The reservoir, it will be the battery} \\

\medskip
Sara (28)     &:	\texttt{Yeah, yeah.} \\

************* & \\

\medskip
Lionel (163) &:	\makecell[tl]{\texttt{... the reservoir what is it? Stores the energy.} \\  \texttt{It is the battery, it is the battery that we put here ...}} \\

************* & \\
\end{tabular}

\noindent
(The students try to put a name to an arrow) 
\smallskip

\noindent
\begin{tabular}{ll}
Fabien (423) &: 	\texttt{... what do we write?} \\

Peggy (424)  &:	   \texttt{It is ... hum ... we write energy, do we? If not ...} \\

Fabien (425) &:	  \texttt{Yeah} \\

\medskip
Peggy (426)   &: 	\texttt{The movement of electrons} \\

************* & \\

Lionel (125)  &:  	\makecell[tl]{\texttt{... But may be we have to draw the arrows to show where} \\ \texttt{the current goes}} \\

Fulvia (126) &: 	\texttt{But we do not know where it goes} \\

Lionel (127)  &:	  \texttt{From the terminal + to the terminal -} \\
\end{tabular}
\begin{center}
\line(1,0){250}
\end{center}

\normalsize
In these dialogs, the problem in need of explanation is twofold. On one hand, students readily match \texttt{reservoir} with \textit{battery}, while, properly speaking, \texttt{reservoir} is a concept in the energy domain under construction and \textit{battery} is a word used in everyday life language independently from considerations for the physics of energy. On the other hand, students associate \texttt{transfer\_of\_energy} with \texttt{electrical\_current}  which belong to two different, and in some sense incommensurate, conceptual domains. How then are these associations between disparate entities belonging to different words possible?

The hypothesis of an underlying always available notional level provides an answer. The idea is that many concepts are denoted using \textit{lexical entities} -- such as ``heat", ``work", ``radiation", ``reservoir"-- that carry with them some kind of flexible knowledge that we call notions. When entities from different domains are compared, the comparison would be done \textbf{at the notional level}, and the matching would occur if, at this level, there is enough overlap between the two attached notions. 

In that way, \texttt{reservoir} and \textit{battery} can be matched on the ground that, at the notional level, both can be empty or filled, both keep some kind of fluid that is often associated with some causal properties. Likewise, at the notional level, \texttt{transfer of energy} and \texttt{electrical current} share attributes like being fluids that circulate and are carriers of causality. Therefore, at the notional level, they can be mistaken one for the other. This lack of differentiation shows in many parts of the dialog where students use indifferently ``energy" and ``current". 

\begin{figure}
  \centering
  \includegraphics[width=0.8\linewidth]{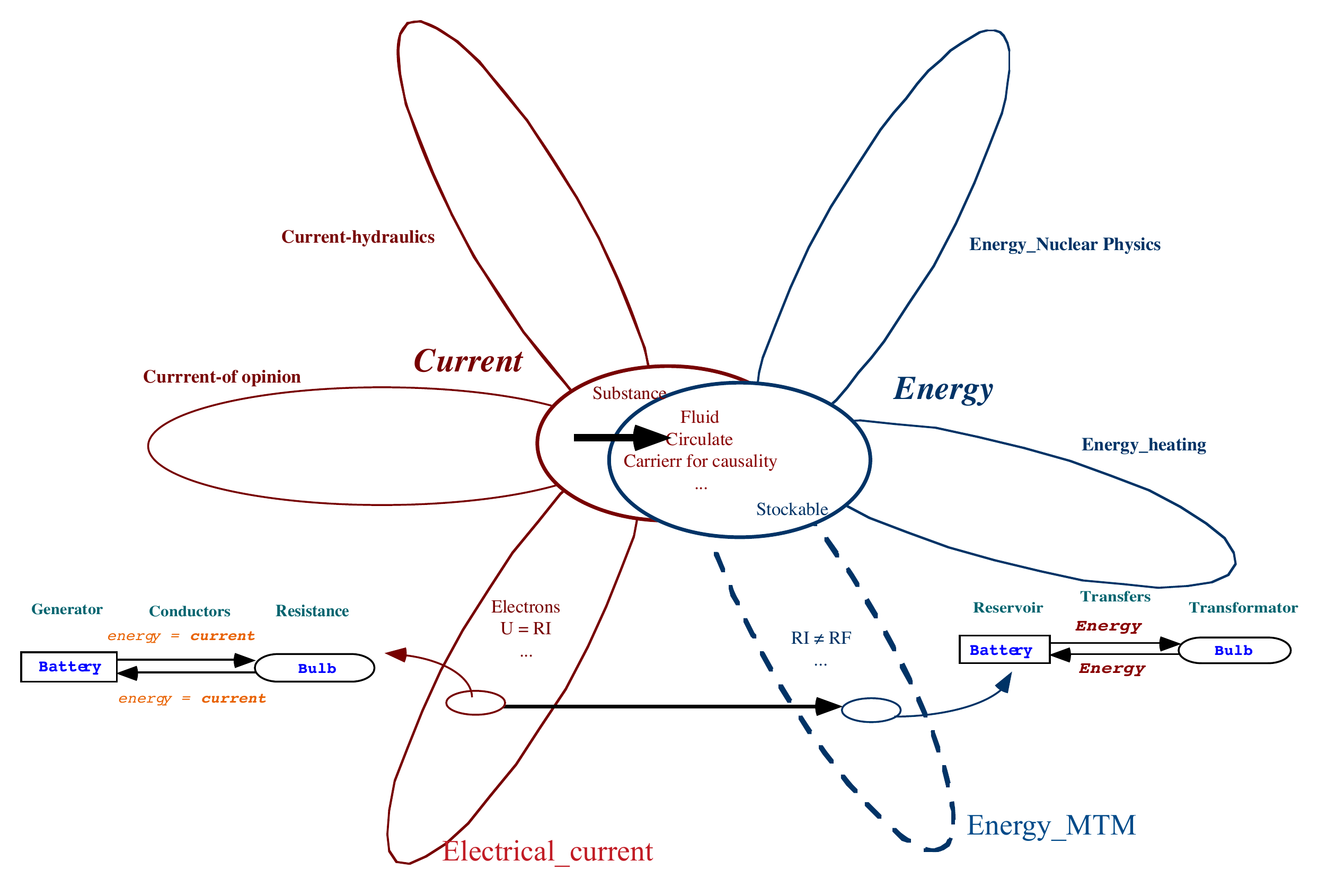}
  \caption{An example of a Janus entity. How the equivalence at the notional level between the notions associated with \textit{energy} and \textit{current} can lead to a confusion of conceptual domains for problem-solving.}  
  \label{fig:figure-notional-1}
\end{figure}

Figure \ref{fig:figure-notional-1} depicts the association in one Janus entity of electrical current and energy considered at the notional level. In the tunnel effect considered throughout this paper, the model of the experimental setting viewed within the conceptual domain of electricity is imported without checking as a valid model viewed within the domain for energy, where it is then reinterpreted. This transposition is automatically done because of the Janus entity of which electrical current and energy are two faces.

It is interesting to note that the linguistic inquiry we performed, as well as the study of current dictionaries, show a sharp disjonction between the scientific concepts and this underlying notional level. They show for instance that “heat” implies the sensation of hotness, which is totally missing in the scientific concept. The notional side of “heat” shows in the following extract, from a dialog between two students.

\begin{center}
\line(1,0){250}
\end{center}
\noindent
\small
\begin{tabular}{ll}
D.	& \texttt{Hum, a bulb that heats a room…} \\

S.	& \texttt{Oh yes, it does!} \\

D.	& \texttt{Such a tiny little bulb?} \\

S.	& \texttt{No, not a little bulb, you are right.} \\
\end{tabular}
\begin{center}
\line(1,0){250}
\end{center}

\smallskip
\normalsize
A physics dictionary for teachers scarcely grants that “a certain amount of heat can sometimes make the temperature to rise”. This raises the question of the learning of concepts from a pre-conceptual stage where only exists a notional layer of knowledge often quite remote from the eventual concepts.

\subsection{An artificial intelligence rendition of notions and concepts}

Many studies, specially in educational sciences, have centered on concepts in complex domains, what they represent, how they function and how they can be learnt (e.g. \citep{reif1992cognition}). For our purpose, it is enough to retain the following description. 

\smallskip
A \textbf{concept} is a node in a relational graph. Leaving aside its semantic content (its links with the world of phenomena), a concept can be entirely defined and considered at this level, that is within the graph. This is what a scientist does when he/she manipulates symbols (e.g. radiation energy, cross section, $Q$, $X$, ...) using relations (e.g. differential equations linking symbols, ...) staying completely at a technical level without any further thought –for a while– for the meaning of the syntactical operations performed. As such, a concept can only be modified through a change of relation(s) in the graph, something which nicely accounts for the discrete nature of conceptual change in contrast with the continuous transformations possible at the notional level.

In addition, a concept gets its semantic content at least partly thanks to a set of recognition procedures that allow to identify positive instances in the world, a set of inference procedures responsible for completion of missing data and prediction, and a store of previously encountered cases, possibly including prototypes.

\textit{If the characterization of the conceptual level is a rather well-marked territory in cognitive science}, this is far from being the case with regard to the notional level. We are therefore very cautious about the following depiction that we offer as a tentative proposal open to criticisms as well as, we hope, positive contributions. 

Our aim here is to provide a characterization for the lack of differentiation between two entities, as well as a possible mechanism for the learning of concepts in connection with the notional level. 

\smallskip
We submit that the notional level consists of a set of schemata that are easy to activate when trying to make sense of the world. For instance:
\begin{itemize}
   \item The notion attached to the term ``energy'' seems to refer to qualities like being a fluid, being a vector for causality, being consumable, and so on.
   \item The notion of \textit{torsion} refers to processes that allow thought experiments, and make possible qualitative predictions. Physicists often mention the kind of mental tinkering they do before any computation in order to get a feeling for what is to be expected with regard to some experiment.
\end{itemize}

\begin{figure}
  \centering
  \includegraphics[width=0.65\linewidth]{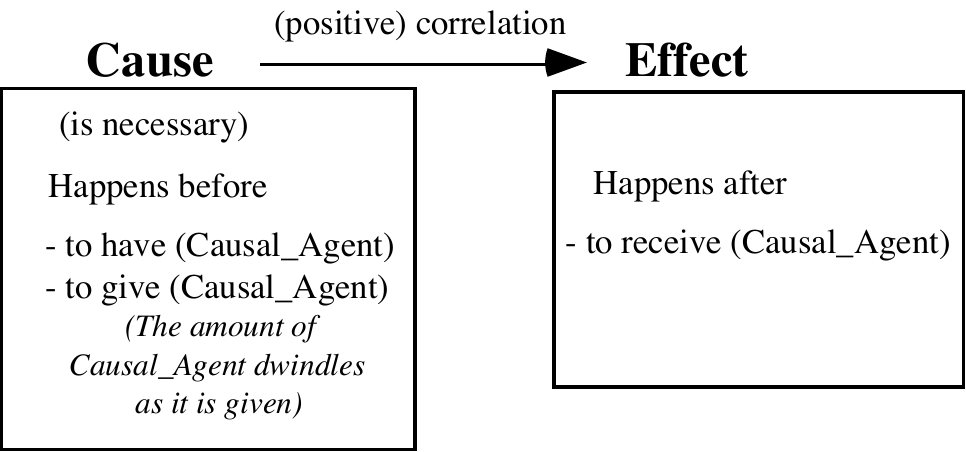}
  \caption{A sketch for the linear causality schema at the notional level.}  
  \label{fig:fig-9}
\end{figure}

The first example above seems to involve mostly properties, while the second one implies inferencing procedures suitable for qualitative reasoning. In our view, notions are schemata that can belong to a very wide scope of nature, ranging from the very abstract, like the schema for linear causality, to the very low-level and concrete, like the schema for transformation (which involves four roles: the \texttt{transformer}, the \texttt{transformee}, the \texttt{initial state} and the \texttt{final state}) (see \citep{collet1996apports} for further details). Like the usual schemata, the notional ones have variables (often of a linguistic nature, that is referred by names like `the cause', `the agent',  `the initial state', and so on). But these variables are associated with loose constraints that allow them to be substituted by a wide range of possible entities with various degrees of fitness and various degrees of necessity. For instance, in the \textit{transform} notional schema (see Figure \ref{fig:fig-10} below), the final state does not have to be identified, and the only constraint is that it is different from the initial state. In addition, the inference procedures, or demons, attached to the variables can be thought of as fuzzy procedures akin to the rules one finds in fuzzy logic, providing only trends and qualitative dependencies between the variables (see for instance Figure \ref{fig:fig-9} below about the linear causality schema). 

In a given context, only a set of the notional schemata, deemed to be relevant, would be considered. For instance, in the energy chain task, some schemata are implied by the seed theory (e.g. \texttt{transformer}, \texttt{reservoir}, ...) and others are activated by the context of a physics class like the one of linear causality. 

At this point, it is interesting to discuss the resemblance and differences between what we call notions and the \textit{Informal Qualitative Models} (IQMs) brought up by \citep{sleeman1989architecture} and further refined notably by \citep{gordon1995modeles}. IQMs were introduced in order to bridge the gap between the data-driven and the theory-driven approaches of scientific discovery computational models. They are basically structural models of actual or hypothetical physical systems that should help to construct qualitative predictive theories or laws which describe or explain the behaviour of a phenomenon in nature. It is our opinion that while the overall organization and functioning of notions and IQMs are similar, there is a distinction as regard to their scope and intent. IQMs are intended to act as kinds of general patterns for the specific laws (generally thought of in terms of numerical dependencies) to be found. As such they are organized in hierarchies and describe the whole phenomenon at hand. In contrast, notions can be of different levels of granularity, sometimes describing local aspects of the phenomena to be interpreted, and sometimes deep underlying principles (as ``the cause is conserved in the effect"). In addition, notions are not primarily intended to provide potential and qualitative dependencies between variables of interest, but are more aimed at supplying explanations in terms of processes taking place in the physical phenomena. All in all, though, IQMs and notions share a lot of assumptions about the role of qualitative and informal knowledge in the scientific discovery process, and their relationship deserves further study.

\section{How tunnel effect activates further adaptation and conceptual learning}
\label{sec-learning}

Two cases must be examined with respect to the opportunities for learning opened when a model has been obtained using tunnel effect:

\begin{enumerate} 
   \item The model obtained remains valid even after being re-interpreted in the target domain under construction.
\vspace{-0.2cm}
   \item The model turns out to be erroneous either when confronted with the world or because internal inconsistencies are discovered within the target interpretation domain.
\end{enumerate}

We study these two cases in turn.

\medskip
\noindent
\textbf{1. The model remains valid}
\smallskip

This is what happened during the construction of thermodynamics by Carnot, Clapeyron, Thomson, Joule, Clausius and others \citep{longair1984theoretical,science-et-vie-1994}. Carnot, influenced by the theory of the caloric (an imponderable fluid with the property of being conserved and which he equated to heat) and by his father's work on the calculation of the efficiency of water mills, devised a cyclic and reversible model describing an ideal steam engine. Thanks to this model, he was able to demonstrate that there exists a maximal efficiency for steam engines, and that it depends on the difference of temperature between the hot source of heat (caloric) and the cold one. Later on, through a series of very meticulous experiments, Joule was able to show that heat was not a conservative quantity and was exchangeable with work. However, it turned out that Carnot's model was in fact neutral with respect to the caloric hypothesis and when re-interpreted in the context of the new theory about heat and work, still remained a very helpful tool for thought experiments, one which eventually lead to the discovery by Clausius of a special state function called entropy. 

We have here one instance of a model obtained through tunnel effect (its cyclic and reversible nature was deeply a result of the belief in the caloric theory even though this was never explicitly expressed by Carnot) which is still valid once the interpretation domain changes. The model by itself cannot therefore act as a trigger for re-evaluation of the target domain, and other symptoms must show. However, because it remains valid, it can help shape the new conceptual system and serve as a test bed for it, potentially through thought experiments as this was the case for Carnot's model in thermodynamics.

It would certainly be interesting to study if some general form of Explanation-Based Learning \citep{mitchell1997machine} could be proposed in order to account for this type of situations. This is as yet an open area for research.

\medskip
\noindent
\textbf{2. The model turns out to be erroneous when re-interpreted}
\smallskip

In our energy chain experiments, this happened either when students realized that the model implied that the energy was flowing back to the battery (which they knew was incorrect), or when they discovered an inconsistency with the target integrity rule stating that the initial energy reservoir should be different from the final one. 

The natural question is then why is the model wrong in the investigated aspect? A re-examination of the path that led to this conclusion in the model can then point towards one of two causes. First, the associations made between entities from the target domain and the source one(s) could be erroneous. For instance, many students question the association they made between electrical current and energy or between the wires and the transfers. This can lead to a differentiation process whereby the target entities gain autonomy with respect to the source ones. Second, the automatic inferencing process that determined the problematic aspect of the model can be disclosed and limitations for its range been set. This is what happened when some students realized that the circular nature of the electrical current did not carry to the energy entity. This inference was henceforth stopped when building a model. 

This short discussion convincingly shows in our opinion that tunnel effects, not only help finding models, even erroneous ones, but that they also provide guidelines for further re-examination and reconceptualisation when needed. This is however an issue that deserves much further work.

\subsection{Learning concepts from notions}
\label{sec-learning-concepts-from-notions}

It is useful to distinguish two stages in the learning of concepts from the notional level. In the first one, the cognitive agent must prove that the faces of the Janus entities are indeed different, i.e. they cannot be identical. The second step then consists in learning a characterization of the schemata that were deceived by Janus entities in order that only one of their face can thereafter, at the conceptual level, satisfy the roles. This is the case for instance when the transformation schema is specialized and constrained so that the initial state and the final one must fulfill an invariance condition (the energy is conserved). At this stage, we are not ready yet to provide a full account of these two steps which involve complex interactions between schemata. We will merely describe one instance of each step in order to illustrate the nature of the mechanisms at play.

\medskip
\noindent
\textbf{1.	How to detect that two entities cannot play the same role: an example}

	Section \ref{sec-analysis-of-the-experiment} showed that most pairs of students did not, at first, differentiate energy from electrical current, and therefore, through tunnel effect,  produced a circular model for the energy circulation. The model thus obtained was clearly then interpreted within the energy domain as testimoned by numerous exchanges (see Section \ref{sec-janus-entities} with examples of exchanges) where the model is interpreted in terms of energy with predictions that the energy will flow back to the battery which, they say, cannot be the case. Since this is the point where a difference between the properties of energy and of electrical current are uncovered, it is interesting to further investigate it. 

Energy is considered by students as a Causal\_Agent. The battery is known to be a reservoir of energy, that is a source of Causal\_Agent, and the state of the lamp as being a kind of Effect. Therefore, according to the linear causality schema, the source of energy must increasingly wear out as it gives away energy. That would not be the case if energy was flowing back to it, which is exactly what the model predicts. Ad absurdio, then, energy cannot flow back to the battery, and therefore it cannot be identical to electrical current which is known, by definition, to return to the battery. 

	We have an example here of a reasoning process involving several schemata at the notional level (linear causality, ad absurdio reasoning, giving, receiving, diminishing, and so on). While it is not really complex, it still requires a not trivial coordination, and a complete and precise artificial intelligence modeling has yet to be realized. It is however displayed naturally and apparently effortlessly by most students. Its only output is to point out the impossibility of maintaining an identification of energy and electrical current at the conceptual level. At this stage, a new and more precise characterization of energy is yet to be learned. We believe the subsequent learning process to act on the schemata that were activated in the previous reasoning, but we so far we lack a better description. The following sub-section shows how learning can occur in a simpler context.

\medskip
\noindent
\textbf{2.	Learning through specialization of schemata: from notions to concepts}
\smallskip

	We describe here how the schema for transformation at the notional level could be specialized to become a schema within the conceptual domain for energy. 

\begin{figure}
  \centering
  \includegraphics[width=0.8\linewidth]{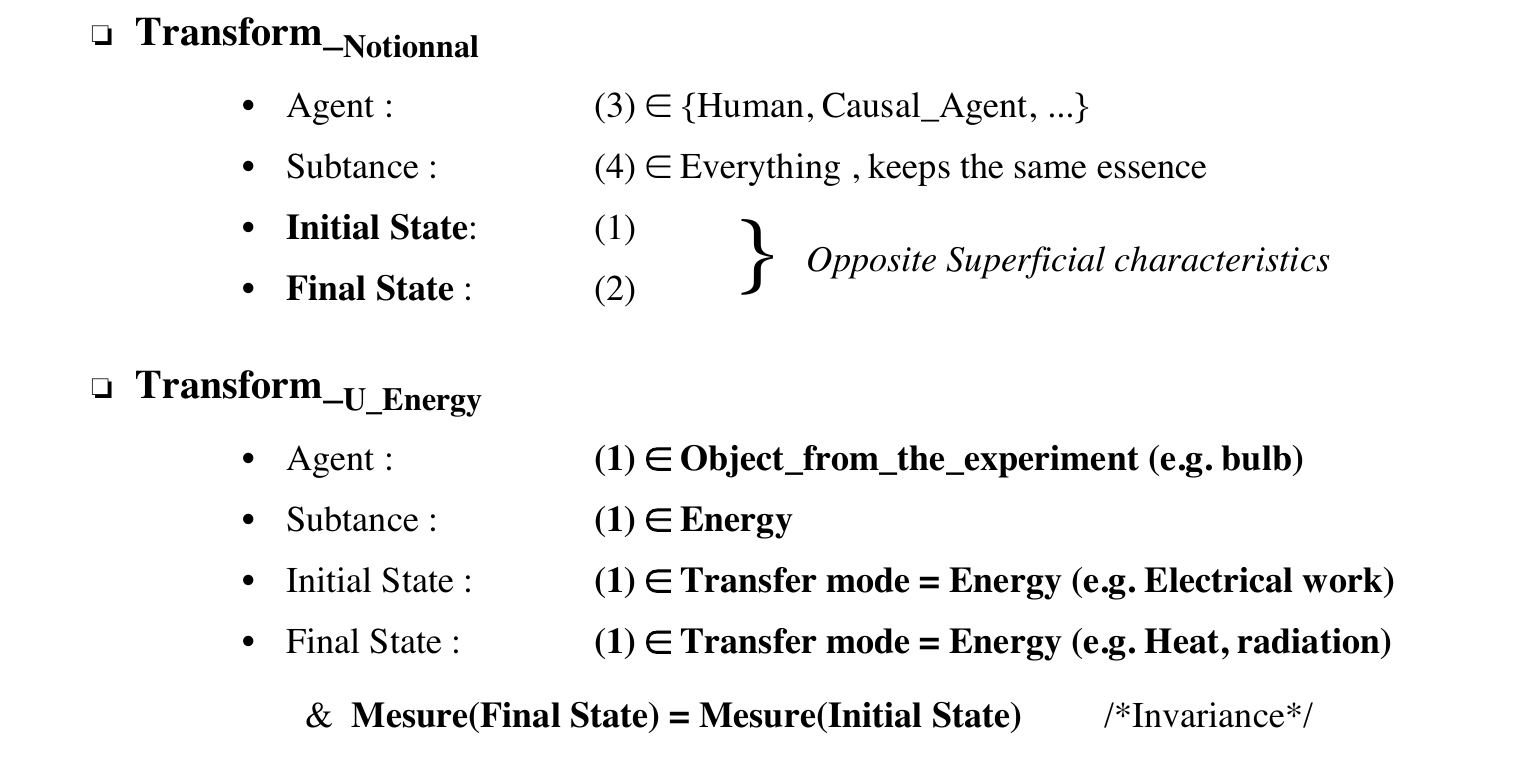}
  \caption{An example of a specialization of a notional description of one entity into a conceptual description obtained through bottom-up generalization of an experimental instance (here the battery-bulb experimental setting in which we have added an hypothetical mean to measure that the energy is the same before and after the transformation, hence the invariance condition).}  
  \label{fig:fig-10}
\end{figure}

	At the notional level, the schema for \texttt{Transform} would involve four roles with rather loose specifications allowing many various arguments to satisfy them. One way to specialize this schema is to take its instanciation in the case of some phenomena as a positive example for the target conceptual schema, and to specialize the constraints attached to each role as tightly as possible to fit the positive instance (bottom-up generalization). This could give a schema such as the one in Figure \ref{fig:fig-10}. 

\subsection{How concepts come to be notions}

Our study of the development of conceptual domains would not be complete if we left aside the puzzling fact that almost every concept of a well-defined conceptual domain is sooner or later deemed to get a notional extension. For instance, the concept of entropy, which is highly technical and counterintuitive in the thermodynamics formulation (involving a state function defined with respect to an ideal Carnot's cycle), has become part of the everyday life stock of expressions (e.g. ``go and get rid of the entropy in your room before asking me for candies"). How is this possible? And how does it fit in our learning scheme? These are questions for further research.

\section{The tunnel effect vs. analogical reasoning}
\label{sec-analogy-vs-tunnel-effect}

Very few inference mechanisms have been proposed that  deal with the transfer of information between different conceptual domains. Analogical reasoning is one of them —the most famous—, blending is another one \citep{fauconnier1998conceptual}, and, we submit, tunnel effect is a contender too. A full comparative study of the three of them would be more than interesting, but is beyond the scope of this paper. However, we believe that a comparison with analogical reasoning might help to enlighten some characteristics of the tunnel effect as an inferencing mechanism. We will concentrate in each case on the conditions for a transfer between interpretation domains to occur, and on the information content that is transferred. 

According to the dominant view on analogy (e.g. \citep{falkenhainer1989structure,greiner1986learning}), analogical reasoning involves the interpretation of two cases, —called the source case for the supposedly well-known one, and the target case for the one to be completed—, that may be interpreted within two different interpretation domains (e.g. the solar system as a source case and the supposedly ill-understood atom system as a target one). Each case is supposed to be represented as a graph of relations and nodes standing for primitive concepts. Analogical reasoning implies then that a best partial match be found between the two graphs, and, in a second step, that the part of the graph representing the source case with no counterpart in the target case representation be copied, translated and added to the target representation in order to fill the missing part. Many questions arise as to the principles that should govern both the matching operation, the translation and the transfer, not to speak about subsequent verification and adaptation. Deep concerns have also been expressed about the interpretation process of the two cases during analogy and the ensuing representation of the cases (e.g. \citep{hofstadter1996fluid,mitchell1993analogy}). It is important to note that both domains —the source and target— must be sufficiently well understood in order that the respective conceptual primitives be identified, put in hierarchy and potentially matched. This view of analogical reasoning thus prevents the consideration of a target domain that would be in gestation and of which conceptual primitives would be very uncertain.  

If we consider then the analogical inferencing mechanism as a kind of black box with inputs and outputs, the inputs consist in the source and target conceptual domains (the conceptual primitives and their relationships (including the said over-important hierarchies) and in the two cases (be they already represented as some would pretend is realistic, or be they interpreted in the context of the analogy as others would insist is unavoidable). The black box then searches for one satisfying matching between the two cases (given as rigid representations or not) and computes the completion of the target case representation. The output or information gained in the operation consists therefore in the added features and properties of the target case.

%

\begin{figure}
  \centering
  \includegraphics[width=1.02\linewidth]{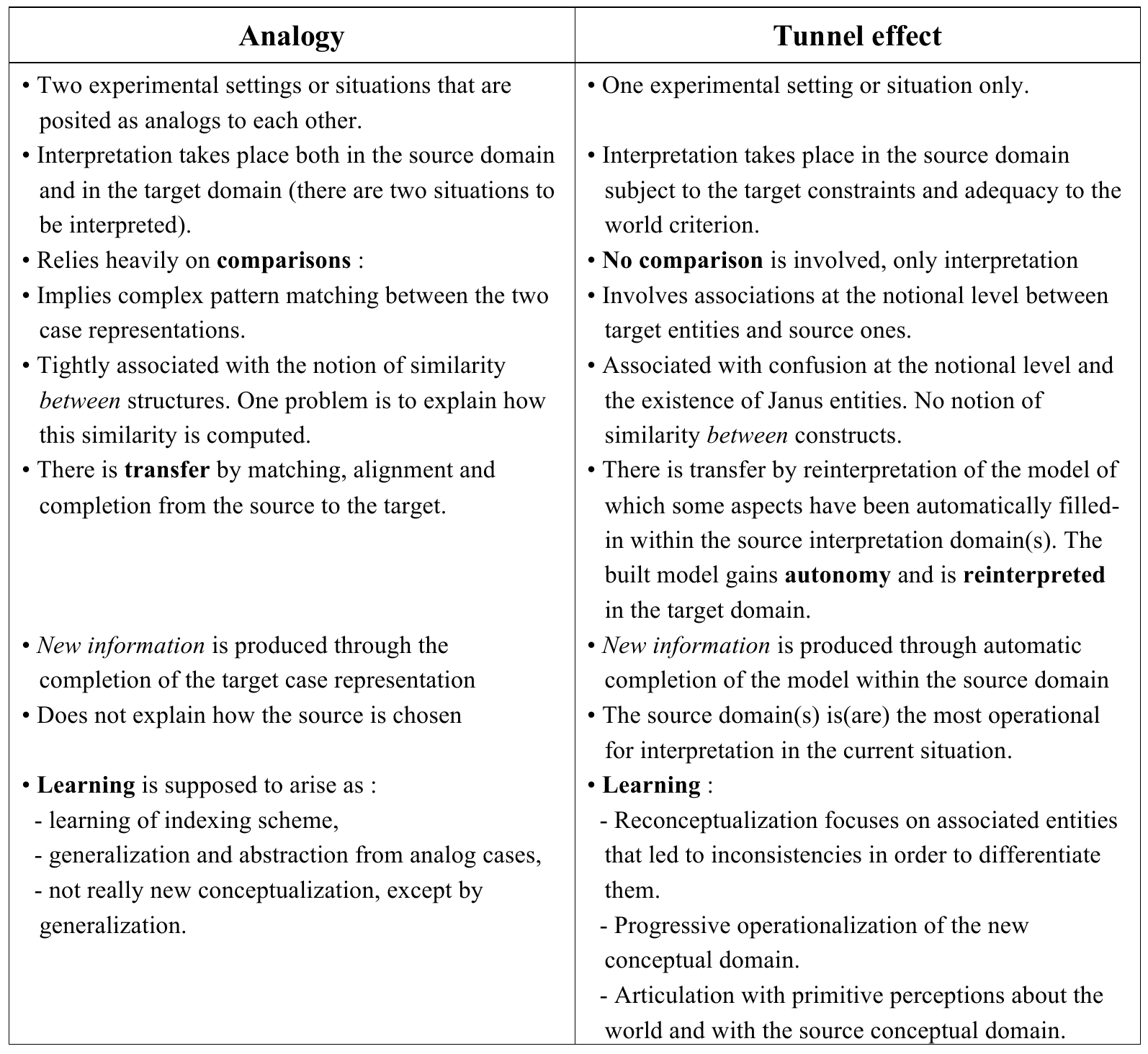}
  \caption{A summary of the main features of analogical reasoning versus features of cognitive tunnel effect.}  
  \label{fig:fig-comparison-analogy-tunnel-effect}
\end{figure}

In contrast, tunnel effect only involves the interpretation of a single situation or case (e.g. an experimental setting or a set of phenomena). The input of the tunnel effect black box consists in the operational source interpretation domain(s), the target criteria that specifies the target interpretation domain (including preconceptions about some target entities, their properties and relationships), and the case (situation or set of phenomena) to be interpreted and understood in the target interpretation domain (e.g. the battery-lamp experiment to be interpreted in terms of energy exchanges, the electromagnetic interactions as measured in Faraday's experiments in terms of a theory in germ in Maxwell's head, or the steam engines in terms of heat and work and other related variables in the nascent thermodynamics). The black box then searches for a model of the case satisfying the target criteria. Because most target entities are not yet operational and interpretable directly in the world, they have to be translated in terms of the more operational interpretation domains given as inputs. In this translation process, submitted to the target criteria, and during model building, some aspects of the model may be automatically filled up through automatic inferencing within the source domain(s) (as is the case when the arrows for transfers are automatically specified when it is decided to translate energy transfer from the notion of electrical current). The output or information gained in the operation consists in the unexpected (because not planned) consequences of the model when interpreted within the target interpretation domain, or in the experimental setting if some target entities are already partially interpretable in the world (as is the case for "energy" for 16-17 years old students).

In both analogical reasoning and tunnel effect, the detection of discrepancies between the resulting model and the world or of other inconsistencies opens opportunities for learning. The difference lies in the fact that tunnel effect is intrinsically intended towards the process of building the domain interpretation domain (through the setting up of connections between this domain, the operational ones in the context and the world) whereas analogical reasoning is oriented towards the completion of some specific case with the help of another 'similar' one. While failed analogies may lead to reconceptualisation in the target interpretation domain, this is much less direct than the learning that may occur when a tunnel effect has produced an unfit model of the world in the interpretation domain (see Figure \ref{fig:fig-comparison-analogy-tunnel-effect} for a summary).

\section{Conclusion}
\label{sec-conclusion}

This paper takes seriously the idea that cognition may imply the existence (and coexistence) of several different interpretation universes, and that a specially important type of learning consists in acquiring new ways of interpreting the world or some aspects of it. In our study we focused on the learning of a new target conceptual domain from the currently operational interpretation one(s) when the attention of the cognitive agent is driven towards the interpretation and understanding of some phenomenon or set of phenomena. 

In studying the type of conceptual learning at play when students are learning a new conceptual domain or when scientists are struggling to find new ways to account for the world, we discovered the important role that one, so far never mentioned, mechanism can play in transferring information from one knowledge domain to another. This mechanism, that we call tunnel effect, facilitates the discovery of interpretations of the world in terms of a new and still ill-known conceptual domain. Its main features are the following:

\begin{enumerate}
   \item It implies a \textit{lack of differentiation between entities} belonging to different domains

   \item While building an expressed model of the phenomena, there might be \textit{illicit transfers fo inference processes from one source entity  to an associated target one}. This eases the construction of models by providing inference mechanisms from the source domain(s) that make up for the as yet non-existent inference mechanisms of the target domain.

   \item If the thus obtained model is then reinterpreted entirely within the target domain, the results of the illicit inferencing processes may bring out \textit{unforeseen consequences} that may satisfy or not the target constraints defining the target domain.

   \item It is then possible that \textit{reconceptualisation occurs}, either to evolve conceptual definitions that help establish differentiation with other concepts and notions, or to make target entities more operational by articulating them more directly to the experimental world.
\end{enumerate}

The essential ingredient in this transfer mechanism lies in the former lack of differentiation between would-be conceptual entities. To account for this, we appealed to the idea of a notional level of knowledge and discourse. While tackling situations in terms of ill-mastered conceptual domains, the natural tendency would be to reason at the notional level. This would allow flexible matching between otherwise ill-known entities. It would explain how scientists may often find their inspiration from other conceptual domains by borrowing concepts through their notional face. This would also provide a basis for learning through specialization. And finally, this may trigger tunnel effect. 

Tunnel effect is an effective and cognitively economical way of transferring knowledge from one conceptual domain to another in construction. Thanks to it, inferencing procedures can be used automatically where they should not if a cautious and systematic conceptual learning was taking place. This facilitates the obtention of candidate interpretations of the world, a process that otherwise could be lengthy and costly. In addition, tunnel effect provides more than a somewhat blind exploration process in that it supplies a focus for attention and reconceptualisation in case the candidate solution turns out not to satisfy the target constraints. Reasoning and learning can then bear on the discovery of distinguishing properties of the erroneously confused  entities. 
Much remains to be done in order to precisely characterize the notional level and the learning that can take place as a result of tunnel effects. However, we believe that our attempt at modeling the learning of new conceptual domains through a mechanism like tunnel effect has some important philosophical implication in that it deeply changes the classical viewpoint according to which learning new conceptual systems must be necessary accompanied by painful and conscious confrontations with some contradiction between the `old' theory and facts. Our approach, by contrast, underlines the constructive role that existing conceptual domains can play in learning a new one. Furthermore, tunnel effect is inherently cognitively cheap. It demands for confusion to take place at the notional level, and for reinterpretation within the target domain. This is easier by far than analogical reasoning, as classically presented in artificial intelligence, which implies costly optimization of complex graph matching. 

Tunnel effect thus provides an economical way of transfering knowledge from one conceptual domain to  another and it offers a focus for subsequent conceptual learning. It seems interesting to investigate further its functioning and its range of application in scientific discovery and in the teaching of science.




\vskip 0.2in
\bibliography{bibliographie}

\end{document}